\documentclass[11pt]{article}

\usepackage[preprint]{acl}

\usepackage{times}
\usepackage{latexsym}

\usepackage[T1]{fontenc}

\usepackage[utf8]{inputenc}
\usepackage{amssymb}
\usepackage{newunicodechar}
\newunicodechar{—}{---}
\newunicodechar{–}{--}
\newunicodechar{…}{\ldots{}}
\newunicodechar{§}{\textsection{}}
\newunicodechar{·}{\ensuremath{\cdot}}
\newunicodechar{×}{\ensuremath{\times}}
\newunicodechar{±}{\ensuremath{\pm}}
\newunicodechar{−}{\ensuremath{-}}
\newunicodechar{→}{\ensuremath{\rightarrow}}
\newunicodechar{↓}{\ensuremath{\downarrow}}
\newunicodechar{↑}{\ensuremath{\uparrow}}
\newunicodechar{⇒}{\ensuremath{\Rightarrow}}
\newunicodechar{≈}{\ensuremath{\approx}}
\newunicodechar{≤}{\ensuremath{\leq}}
\newunicodechar{≥}{\ensuremath{\geq}}
\newunicodechar{∈}{\ensuremath{\in}}
\newunicodechar{∪}{\ensuremath{\cup}}
\newunicodechar{∖}{\ensuremath{\setminus}}
\newunicodechar{⌈}{\ensuremath{\lceil}}
\newunicodechar{⌉}{\ensuremath{\rceil}}
\newunicodechar{★}{\ensuremath{\bigstar}}
\newunicodechar{Δ}{\ensuremath{\Delta}}
\newunicodechar{ρ}{\ensuremath{\rho}}
\newunicodechar{σ}{\ensuremath{\sigma}}

\usepackage{microtype}

\usepackage{inconsolata}

\usepackage{graphicx}

\usepackage{booktabs}

\usepackage{float}
\usepackage{tcolorbox}
\tcbuselibrary{breakable}
\definecolor{UserColor}{rgb}{0.6,0.6,0.6}
\newtcolorbox{myblock}[1]{colback=UserColor!10, colframe=UserColor, coltitle=white, title=#1, fonttitle=\bfseries, breakable}

\usepackage{amsmath}
\usepackage{colortbl}
\usepackage{multirow}
\usepackage{array}

\title{An Axiomatic Benchmark for Evaluation of Scientific Novelty Metrics}

\author{Miri Liu \& ChengXiang Zhai \\
Department of Computer Science\\
University of Illinois, Urbana-Champaign (UIUC)\\
\texttt{\{miri3, czhai\}@illinois.edu}
}

\begin{document}
\maketitle
\begin{abstract}
The rigorous evaluation of the novelty of a scientific paper is, even for human scientists, a challenging task. With the increasing interest in AI scientists, it is becoming more and more important that this task be automatable and reliable, lest attention and compute be wasted on ideas that have already been explored. Due to the challenge of quantifying ground-truth novelty, however, existing novelty metrics generally validate against noisy, confounded signals such as citation counts or peer review scores. We introduce a benchmark that compares novelty metrics without requiring explicit novelty labels. It tests whether scores move correctly under controlled pool manipulations, organized under three axioms requiring that scores fall as the pool covers more of a paper's content, rise as the pool loses relevance, and fall as the pool moves later in time. Indeed, we show that even for the most direct human signal, ICLR reviewer novelty scores, the axis of novelty is entangled with quality. Across ten systems, from embedding metrics to AI scientist novelty checks, we find that surface redundancy is largely solved but conceptual redundancy is not, and that embedding metrics and LLM-based metrics both have their places; future metrics should combine such approaches for better novelty evaluation. We release our benchmark and evaluation code to enable this research.
\end{abstract}

\section{Introduction}
Novelty is a foundational expectation of scientific work — whatever else a paper contributes, it must advance the state of knowledge in some real way. As more and more papers are written every year \citep{bornmann2021growthratesmodernscience, Hanson_2024} and research communities grow more siloed \citep{park2023papers, gates2025increasingfragmentationglobalscience, evans2008}, it becomes impossible for scientists to meaningfully keep up with and evaluate the literature. A reliable, automated novelty metric would cut through this noise, surfacing genuinely novel work.

Such a metric is also increasingly desirable given the recent explosion of work on AI for scientific discovery. These works generally treat novelty only lightly, such as by using human annotation of a subset of generated ideas \citep{si2024llmsgeneratenovelresearch, yang2024largelanguagemodelsautomated} or relying on LLMs as judges of relative novelty \citep{ghareeb2025robinmultiagentautomatingscientific, li2024chainideasrevolutionizingresearch, lu2024aiscientistfullyautomated, baek2025researchagentiterativeresearchidea}. However, follow-up work has already found that self- and human-assessed novelty of LLM-generated research ideas is systematically inflated \citep{si2025ideationexecutiongapexecutionoutcomes}. Without reliable novelty metrics, AI scientist pipelines could end up optimizing for something that only \textit{looks} novel.

This raises the question of how to actually evaluate a scientific novelty metric. Novelty is a slippery quality; existing evaluations either correlate against human annotations --- survey responses, review scores, or prize designations \citep{semnovel, fasttext, rnd, novascore} --- or against automatic proxies such as citation count \citep{semnovel, luo2022combination}. These signals have clear limitations. Human annotation is expensive, varies with the annotator's knowledge and judgment, and, as we show in \S\ref{sec:motivating}, is influenced by a paper's overall quality. Meanwhile, citation counts conflate novelty with impact, and LLM-as-a-judge replacements carry documented biases \citep{shi-etal-2025-judging,spiliopoulou2025playfavoritesstatisticalmethod,panickssery2024llmevaluatorsrecognizefavor,ye2024justiceprejudicequantifyingbiases}.

Taking this into consideration, how can we objectively compare multiple novelty metrics, understand their relative strengths, improve them, and perhaps combine them? Our solution draws on axiomatic thinking, previously applied to elusive qualities in retrieval evaluation \citep{fang_informationretrieval}: we define a set of axioms that would apply to any reasonable scientific novelty metric. The extent to which a metric satisfies the axioms then gives a clear behavioral profile.

We make three contributions. \textbf{(1)} We introduce the \textbf{first axiomatic benchmark for scientific paper novelty metrics}, defining three axioms instantiated by behavioral probes, on ten tasks spanning three domains of AI research; it is metric-agnostic and easily extensible. \textbf{(2)} We \textbf{analyze a broad range of existing novelty systems}, showing that none performs satisfactorily across the board, that the systems fail in actionable ways, and that nearly every probe is satisfied by \textit{some} existing system. \textbf{(3)} We show empirically that \textbf{the commonly used correlation with reviewer novelty scores is problematic} despite seemingly being the ideal validation strategy for novelty metrics; the labels themselves are highly confounded with quality.

\section{Related Work and Taxonomy}
\label{sec:related_works}

\subsection{Novelty Metrics and Systems}
The rise of AI for scientific discovery has renewed interest in the problem of how to quantify scientific novelty. \cite{zhao2025novelty} categorizes the methodology of scientific novelty metrics into those that draw on citation data, those that draw on textual data, and those that draw on multiple sources of data. Citation-based metrics of novelty \citep{uzzi2013} or similar concepts such as disruptiveness \citep{wu2019large} analyze citation counts, patterns, and relationships. Textual data metrics, which are the focus of this paper, can include keyword- or entity-based metrics \citep{mishra2016novelty, ruan2025}; sentence-based or contribution-based metrics, including topic modeling approaches \citep{wang2024effective, sendhilkumar2013novelty}; and embedding-based metrics \citep{yin, rnd, fasttext, semnovel}. 

Beyond scientific literature, novelty detection has traditionally been studied in IR and NLP at the sentence and document level \citep{soboroff-harman-2005-novelty, ghosal-etal-2018-novelty, novascore, ghosal2022novelty}, though generally on documents far smaller than scientific papers.

Researchers have also proposed a range of LLM-based novelty judgments over papers \citep{zhang2026opennoveltyllmpoweredagenticverifiable, lin2024evaluatingenhancinglargelanguage} and research ideas \citep{si2024llmsgeneratenovelresearch, qiao2026innoevalresearchideaevaluation}. Many more automatic novelty judgments are embedded implicitly or explicitly in autonomous research pipelines, where they gate which ideas are pursued. Some checks consult retrieved literature or records of previously seen ideas \citep{lu2024aiscientistfullyautomated, li2024chainideasrevolutionizingresearch, liu2026autoresearchclawselfreinforcingautonomousresearch}; many more are self-referential, meaning a reviewer agent may make use of parametric knowledge or simply rank generated ideas against each other \citep{baek2025researchagentiterativeresearchidea, schmidgall2025agentlaboratoryusingllm, weng2025cycleresearcherimprovingautomatedresearch, tang2025airesearcherautonomousscientificinnovation, keya2025sciideacontextawarescientificideation, weng2025deepscientistadvancingfrontierpushingscientific, ghareeb2025robinmultiagentautomatingscientific}. Our framework aims to help audit and ultimately improve these novelty evaluation components.

\subsection{Validation and Axiomatic Evaluation}
Existing validation efforts and frameworks for novelty metrics rely on the same problematic external signals discussed above, such as prizes, expert opinion, and citation behavior \citep{yin, fasttext, semnovel, wu2019large, shibayama2021novelty, wang2024effective, fontana2020new, bornmann2019novelty, amplayo2019evaluating}.

Axiomatic frameworks have been productively applied in adjacent evaluation contexts. In information retrieval, \citet{fang_informationretrieval} introduced a set of formal constraints that retrieval functions should satisfy, and further work expanded on these to identify axioms for evaluating effectiveness metrics \citep{busin2013}, classification metrics \citep{sebastiani2015}, and document organization tasks \citep{amigo2013}. In software engineering, metamorphic testing similarly uses necessary relations between outputs on systematically transformed inputs to test correctness where no reliable oracle exists \citep{chen1998metamorphic}.

To our knowledge, however, there is no analogous work for novelty metrics. The closest existing works are NovBench, which assesses the text quality of LLM-generated novelty assessments \citep{wu2026novbenchevaluatinglargelanguage}, and RINoBench, which benchmarks LLM judgments of research-idea novelty against gold labels aggregated from ICLR 2022--23 reviewer novelty ratings \citep{schopf2026ideanovelautomatedbenchmark}. Unlike these works and related critiques of LLM performance on novelty assessment \citep{jiang2026hindsightevaluatingllmgeneratedresearch, sinhahajari2026limitsllmasjudgescientificnovelty, beel2025evaluatingaiscientist}, we depend on no human-derived novelty labels; rather, we determine whether novelty scores move correctly under controlled pool manipulations.
\section{Motivating Study}
\label{sec:motivating}

Perhaps the ideal way to validate a scientific novelty metric would be to collect human novelty scores; this is the approach taken by the recent Metascience Novelty Indicators Challenge,\footnote{\url{https://challengeworks.org/challenge-prizes/metascience-novelty-indicators/}} for instance. The most abundant publicly available similar data is review scores from OpenReview venues, especially ICLR, which in 2022 and 2023 included fields explicitly referring to novelty --- \texttt{technical\_novelty\_and\_significance} and \texttt{empirical\_novelty\_and\_significance}. This data is the most direct human novelty signal available at scale, since reviewers are explicitly asked to rate novelty. However, as we show in this section, even these novelty scores are in practice inseparable from overall \textit{execution quality}. Correlation with the reviewer-assigned scores, then, is not necessarily useful for understanding the actual novelty-based performance of a novelty metric.

To test this and evaluate the usefulness of this validation data, we construct two $n{=}200$ samples of papers from ICLR 2022 and 2023; for each of $\{\text{technical novelty}, \text{empirical novelty}\}$, we take the 100 highest- and 100 lowest-rated papers with high reviewer consensus. We build a reference pool for each paper, score the papers with the systems highlighted in \S\ref{sec:taxonomy}, and compute Spearman's $\rho$ against the reviewer field mean as well as the ROC AUC for recovering a paper's high/low bucket (construction details in Appendix~\ref{app:motivating}). As Table~\ref{tab:iclr} (Appendix~\ref{app:motivating}) shows, every system except one performs poorly on both dimensions ($|\rho| \leq 0.18$, AUC 0.42--0.62). This happens even though our problem setting is easy by construction, since we have filtered not only for extremes but for high-agreement extremes. Several systems in fact correlate negatively --- significantly so for AutoResearchClaw (ARC) --- and the AI Scientist judge, the only other system with a positive signal, reaches only $\rho = +0.18$ (technical). The exception is RAG-Novelty, at $\rho = +0.69$ (technical) and $+0.58$ (empirical), with bucket AUCs of 0.89 and 0.83.

This particular success is not evidence of good novelty assessment. Our consensus filtering makes the novelty buckets nearly identical to acceptance (bucket and decision agree for 93.5\% (technical) and 97.0\% (empirical) of papers); when all reviewers agree a paper is low-novelty, the paper is essentially always rejected. We find that RAG-Novelty's score predicts the accept/reject decision at AUC 0.83. When restricting to accepted papers only (and thus removing the acceptance axis), every correlation collapses to null, including RAG-Novelty's ($|\rho| \leq 0.15$, all $p > 0.15$, $n{=}89$ and $96$).

Additionally, rescoring every paper with its retrieval summary swapped for those of a \textit{random} other paper leaves the correlation unchanged ($\rho = +0.68$ technical, $+0.58$ empirical). We claim this shows RAG-Novelty's rating here is primarily the LLM's read of the title and abstract, especially because RAG-Novelty fails later probes like identifying a verbatim copy of the evaluated paper in the reference pool (\S\ref{sec:results}). This quality signal also survives deep paraphrase (see Appendix~\ref{app:motivating}), making it challenging to remove.

A good novelty metric may well succeed on this data, since quality and novelty travel together in human research work. We argue, however, that this validation data is \textit{uninformative} for understanding where existing metrics fail (or even succeed, as with RAG-Novelty), and therefore for understanding how to improve them. To solve this problem, we introduce specific principles that we argue reflect the human understanding of novelty, and probes that test metric agreement with these principles in \S\ref{sec:framework}. The axiomatic framework is meant to enhance the current way of validating novelty metrics, not to replace it.

\section{Framework}
\label{sec:framework}

Two principles underlie our framework. First, scientific papers do not have \textit{intrinsic} novelty; a paper is novel only relative to some body of prior work. Accordingly, every test constrains the relationship between two scores assigned to the \textit{same} paper under a controlled change to its reference pool, rather than the absolute novelty score itself. Second, each test is a \textbf{consistency property} expected to hold \textit{in aggregate}, not a guarantee for every individual paper. We therefore evaluate each as a pass rate over many papers and make claims only at the level of \textbf{systems}.

We organize the tests as three \textbf{axioms} --- monotonicity principles over pool transformations --- each instantiated by a family of concrete \textbf{probes}. We treat the axioms as necessary but not sufficient conditions: a system may pass individual probes without being a robust measure of novelty (\S\ref{sec:p1results}), but a system that fails them is not adequately measuring novelty. Throughout, $P$ is a focal paper, $C$ its base reference pool (prior work on $P$'s task published before $P$; see \S\ref{sec:data_collection}), and $s(P, \cdot)$ the evaluated system's novelty score for $P$ against a given pool.

\subsection{Axiom R --- Redundancy}
\textit{The novelty score must decrease as the pool's coverage of $P$'s content increases.} 

\paragraph{R-exact:} $s(P,\, C \cup \{P\}) < s(P,\, C)$. If an exact copy of $P$ is added to the reference pool, $P$ is non-novel.

\paragraph{R-paraphrase:} $s(P,\, C \cup \{\widetilde{P}\}) < s(P,\, C)$. If a paraphrased $\widetilde{P}$ is added instead, $P$ should still score lower, as its content is still present.

\paragraph{R-partial($f$):} $s(P,\, C_{f}) < s(P,\, C)$ for each $f \in \{25, 50, 75, 100\}\%$, with $s(P,\, C_{f})$ decreasing in $f$. $C_f$ plants a fraction $f$ of $P$'s atomic claims, \textit{paraphrased}, each into a distinct pool paper. Because some of $P$'s claims already appear in the pool, $P$ is progressively less novel as more claims appear.

\paragraph{R-verbatim:} $s(P,\, C^{\mathrm{verb}}_{100}) < s(P,\, C)$. The same plant as above, but with the claims kept \textit{verbatim}.

\paragraph{R-concentrated:} $s(P,\, C^{\mathrm{conc}}_{100}) < s(P,\, C^{\mathrm{dist}}_{100}) < s(P,\, C)$. We approximate a duplicate of $P$ by concentrating all planted claims into a single pool paper; this should score lower than distributed coverage.

\subsection{Axiom V --- Relevance}
\textit{The novelty score must increase as the pool becomes less relevant to $P$.}

\paragraph{V-far:} $s(P,\, C_{\mathrm{far}}) > s(P,\, C)$. $C_{\mathrm{far}}$ is a pre-$P$ pool from a task in a different domain (Table~\ref{tab:tasks}), against which $P$ should appear more novel than against its own field.

\paragraph{V-near:} $s(P,\, C_{\mathrm{near}}) > s(P,\, C)$; \quad \textbf{V-graded:} $s(P,\, C_{\mathrm{far}}) > s(P,\, C_{\mathrm{near}})$. $C_{\mathrm{near}}$ is a pool from a different task in the same domain, so it sits between $C$ and $C_{\mathrm{far}}$ in relevance, and the scores should order accordingly.

\paragraph{V-cited:} $s(P,\, C \setminus \mathrm{cit}) > s(P,\, C)$. $\mathrm{cit}$ is the set of papers $P$ itself cites, the work it is in dialogue with; removing them should make $P$ appear more novel.

\paragraph{V-primacy:} $s(P,\, \mathrm{cit}) < s(P,\, C \setminus \mathrm{cit})$. When judged against only its own declared prior art, which should be extremely relevant, $P$ should appear less novel.

\subsection{Axiom T --- Temporality}
\textit{The novelty score must decrease as the pool's time window moves later.}

\paragraph{T-accumulation:} $s(P, W_1) > s(P, W_2) > s(P, W_3) > s(P, W_4)$. $W_1$ (oldest) through $W_4$ (newest) are \textit{equal-count} date quartiles of $C$, so the windows differ only in temporal identity. Against an older window, ideas from the intervening years are absent and $P$ should appear more novel.

\paragraph{T-subsumption:} $s(P,\, W_{+}) < s(P,\, W_4)$ \ and \ $s(P,\, W_{+}) < s(P,\, C)$. $W_{+}$ is an equal-count slice of papers published \textit{after} $P$, disjoint from the base pool. As later work will partially or fully absorb $P$'s contribution, $P$ should appear less novel.

\subsection{Evaluation and Statistical Treatment}
\label{sec:stats}

We evaluate each probe per focal paper as a binary ordering check, then report the total fraction of focal papers that satisfy this check. We test whether each system performs \textbf{above chance} using one-sided binomial tests with Holm correction applied separately for each system; the chance floors are listed with the operational details in Appendix~\ref{app:probes}. For the coverage fractions and temporal windows, which are ordered, we also report Page trend tests. Finally, all probes rest on operational premises, which we show are true by construction or measured (Appendix~\ref{app:premises}); objections to a premise are not objections to the axioms.

The axiomatic framework has three benefits. One, it provides a multi-dimensional, interpretable profile for assessing a novelty metric, which is more informative than a single correlation score; this deepens our understanding of any given metric and provides insight into how to improve it. Two, it provides an objective and inexpensive way to make measurements; it avoids the variance of human annotation and can be applied to any new system at any time, whereas human ratings age and must eventually be recollected. Three, it enables monotonic growth of our understanding of desired novelty metric behavior, with the proposed axioms as a baseline theoretical framework that future work can and should expand.
Next, we present experiments demonstrating these benefits through a detailed comparative analysis of the major existing novelty metrics.

\section{Experimental Setup}
\subsection{Data Collection}
\label{sec:data_collection}
To obtain collections of papers that can be reasonably assumed to share a research focus while remaining small enough for tractable pool manipulation, we draw upon the PapersWithCode archive available on Hugging Face.\footnote{https://huggingface.co/datasets/pwc-archive/papers-with-abstracts} The archive (dated July 29 2025) includes arXiv IDs, titles, abstracts, and author-submitted task tags; all papers are English-language. After normalizing task titles to resolve duplicate names (Appendix~\ref{app:probes}), we select tasks across three domains: NLP, computer vision, and biomedical AI. All selected tasks are AI research tasks, which is an inherent limitation of the repository; however, since the novelty evaluation of AI-generated research ideas is a primary motivation of this work, AI research is likely an appropriate test bed regardless. We select relatively specific tasks with roughly 1,500 tagged papers each, attempting diversity within each domain. For each task, we also select a ``distant task'' from one of the tasks in the other two domains. We report selected tasks, designated distant tasks, and pool sizes in Table~\ref{tab:tasks}.

\subsection{Evaluated Novelty Systems}
\label{sec:taxonomy}
We categorize the novelty judgments of the works surveyed in \S\ref{sec:related_works} along three axes, the \textbf{output granularity} (free text through binary and ordinal verdicts to a continuous score), the \textbf{pool observability} (explicit corpus, open retrieval, or parametric knowledge), and the \textbf{comparison unit} (whole papers, individual claims, or candidate ideas); Table~\ref{tab:taxonomy} (Appendix~\ref{app:other_systems}) organizes the systems along these axes.

\label{sec:systems}

Of the systems organized in Table~\ref{tab:taxonomy}, we evaluate ten. Eight produce some kind of score or verdict and are thus evaluated under the full suite. Four of these are corpus-conditioned continuous novelty metrics operating on embeddings: the minimum-distance metric of \citet{yin}, relative neighborhood density \citep{rnd}, the t-SNE-based SemNovel \citep{semnovel}, and the local-outlier-factor metric of \citet{fasttext}. NovaScore \citep{novascore}, in the NLP novelty detection lineage, uses LLMs to extract atomic content units, then compares a given document's units against the retrieved pool with entailment. The remaining systems are more recent and developed for AI for scientific discovery. The AI Scientist judge \citep{lu2024aiscientistfullyautomated} is a binary check with literature search; RAG-Novelty \citep{lin2024evaluatingenhancinglargelanguage} rates novelty based on an LLM reading of the evaluated paper along with retrieved neighbors; and ARC \citep{liu2026autoresearchclawselfreinforcingautonomousresearch} is a lexical novelty gate used in AutoResearchClaw. The AI Scientist judge is run on a three-task subset for cost (Appendix~\ref{app:ledger}).

For the last two evaluated systems, which do not output scores, we test the redundancy axiom specifically through scoped adapted readouts. \citet{si2024llmsgeneratenovelresearch} use a pairwise ranking protocol, so we measure the displacement of the focal paper's rank under pool insertions. OpenNovelty \citep{zhang2026opennoveltyllmpoweredagenticverifiable} outputs free-text assessments with cited evidence, so we measure whether it correctly attributes non-novelty to planted prior work. For more details, please see Appendix~\ref{app:other_systems}.

\subsection{Evaluation Protocol}

For each task, we randomly sample 100 focal papers. For each focal paper, a base pool is constructed from all papers tagged with the same task and published strictly before the focal paper's publication date. We use the Semantic Scholar API \citep{Kinney2023TheSS} to retrieve references for each focal paper, which are also added to appropriate pools based on publication year. We fix the probes' operational choices uniformly across systems. Planted claims go to pool papers chosen by TF-IDF cosine similarity, which we use in part because no evaluated system relies on it, and probes are skipped when a pool falls below fixed size thresholds, listed with the remaining operational details in Appendix~\ref{app:probes}. Because \cite{fasttext} embeds only titles, it is excluded from the claim-planting probes.

For systems that output, say, binary verdicts rather than continuous scores, we adapt the readout while leaving the judgment computation untouched. Where the judge model exposes token log-probabilities, we relax the binary verdict to $P(\text{novel})$, using \texttt{gpt-4o-mini} since current model APIs usually refuse log-probability requests (Appendix~\ref{app:ledger}).

Additionally, to control reference pools for systems that retrieve from tools like Semantic Scholar, we sandbox the systems and redirect retrieval to constructed pools. LLM judges remain only weakly sandboxed, since we cannot edit parametric knowledge (see Limitations). Finally, for systems with cached intermediate representations, manipulated content is composed into the representation so pools differ only by the manipulation. For detail on configuration and implementation, see Appendix~\ref{app:adapters}.

\section{Results}
\label{sec:results}

We evaluate the novelty systems against the axioms and probes above. The full results of that evaluation are given in Table~\ref{tab:main}, with per-task tables in Appendix~\ref{app:tables}. Ties count as failures; Table~\ref{tab:ties} decomposes every cell into pass, tie, and wrong-way movement, and Appendix~\ref{app:attribution} analyzes scale effects.

\begin{table*}[t]
\centering
\small
\begingroup
\footnotesize
\setlength{\tabcolsep}{3.5pt}
\begin{tabular}{lcccccccc}
\toprule
\rowcolor{gray!15}
\textbf{Probe} & \textbf{Yin} & \textbf{RND} & \textbf{SemNovel} & \textbf{FastTextLOF} & \textbf{NovaScore} & \textbf{AI-Sci Judge$^{*}$} & \textbf{RAG-Novelty} & \textbf{ARC} \\
\midrule
\multicolumn{9}{l}{\emph{Axiom R --- Redundancy}} \\
R-exact & \textbf{1.00} & \textbf{0.99} & 0.43 & 0.52 & \textbf{0.73} & \textbf{0.70} & 0.28 & \textbf{1.00} \\
R-paraphrase & \textbf{0.99} & \textbf{0.97} & 0.40 & 0.50 & 0.49 & \textbf{0.73} & 0.26 & \textbf{0.60} \\
R-verbatim & \textbf{0.91} & \textbf{0.88} & 0.46 & — & \textbf{0.56} & 0.50 & 0.25 & 0.12 \\
R-partial (25\%) & \textbf{0.66} & 0.53 & 0.51 & — & 0.43 & 0.50 & 0.24 & 0.02 \\
R-partial gradient$^\dagger$ & 0.00 & 0.08 & 0.04 & — & 0.03 & 0.04 & 0.00 & 0.00 \\
R-concentrated $<$ base & \textbf{0.79} & \textbf{0.60} & 0.46 & — & 0.51 & \textbf{0.59} & 0.23 & 0.12 \\
R-concentrated $<$ distributed & \textbf{0.69} & 0.07 & 0.49 & — & 0.35 & 0.54 & 0.18 & 0.12 \\
\midrule
\multicolumn{9}{l}{\emph{Axiom V --- Relevance}} \\
V-far & \textbf{1.00} & \textbf{0.99} & 0.05 & \textbf{0.65} & 0.46 & \textbf{0.89} & 0.29 & 0.03 \\
V-near & \textbf{0.98} & \textbf{0.93} & 0.02 & \textbf{0.59} & 0.43 & \textbf{0.84} & 0.29 & 0.03 \\
V-graded (far $>$ near) & \textbf{0.90} & \textbf{0.59} & 0.50 & \textbf{0.55} & 0.32 & \textbf{0.61} & 0.19 & 0.00 \\
V-cited & \textbf{0.59} & \textbf{0.79} & 0.43 & 0.45 & 0.38 & \textbf{0.64} & 0.23 & 0.03 \\
V-primacy & \textbf{0.59} & \textbf{0.98} & 0.97$^{\mathsection}$ & \textbf{0.62} & 0.33 & \textbf{0.62} & 0.19 & 0.04 \\
\midrule
\multicolumn{9}{l}{\emph{Axiom T --- Temporality}} \\
T-acc.\ $W_1 > W_2$ & \textbf{0.73} & \textbf{0.59} & \textbf{0.58} & 0.51 & 0.31 & \textbf{0.60} & 0.27 & 0.00 \\
T-acc.\ $W_2 > W_3$ & \textbf{0.66} & \textbf{0.57} & \textbf{0.58} & 0.48 & 0.31 & \textbf{0.63} & 0.27 & 0.01 \\
T-acc.\ $W_3 > W_4$ & \textbf{0.60} & 0.52 & 0.53 & 0.48 & 0.32 & 0.54 & 0.21 & 0.02 \\
T-acc.\ monotone$^\dagger$ & \textbf{0.24} & \textbf{0.16} & 0.10 & 0.04 & 0.01 & 0.09 & 0.00 & 0.00 \\
T-sub.\ $< W_4$ & 0.42 & 0.36 & 0.54 & 0.50 & 0.31 & 0.40 & 0.17 & 0.02 \\
T-sub.\ $<$ base & 0.25 & \textbf{0.57} & 1.00$^{\mathsection}$ & 0.55 & 0.29 & 0.40 & 0.19 & 0.02 \\
\bottomrule
\end{tabular}
\endgroup
\caption{Macro pass rates over the 10 tasks (mean of per-task rates) for all systems across the redundancy, relevance, and temporality probes. \textbf{Bold} = pooled one-sided binomial significant at Holm-corrected $\alpha=.05$; $\dagger$ = chained check (chance 0.125--0.25); $*$ = 3-task subset; $\mathsection$ = significant but attributed to a frame artifact (Appendix~\ref{app:attribution}). Cells are single-run pass rates; 95\% binomial intervals are within $\pm 0.03$ ($\pm 0.06$ for the subset).}
\label{tab:main}
\end{table*}

\subsection{Redundancy}
\label{sec:p1results} 
From the redundancy probes we see that while surface redundancy is largely solved, conceptual redundancy is not. Yin and RND can detect exact copies and paraphrases almost perfectly, which is expected from the embeddings distance approach. However, the same design makes claim-level coverage difficult; planted claims sit in different pool papers, meaning no single pool paper will strongly resemble the focal paper, and they reach only 0.66 and 0.53 at 25\% coverage. The two metrics also disagree on the concentrated coverage probe --- Yin scores 0.69, as the minimum embedding distance treats the claim-saturated paper as a near duplicate, while RND scores only 0.07, because a density rank responds more to several moderately close papers. 

\begin{figure}[t]
\centering
\includegraphics[width=0.92\columnwidth]{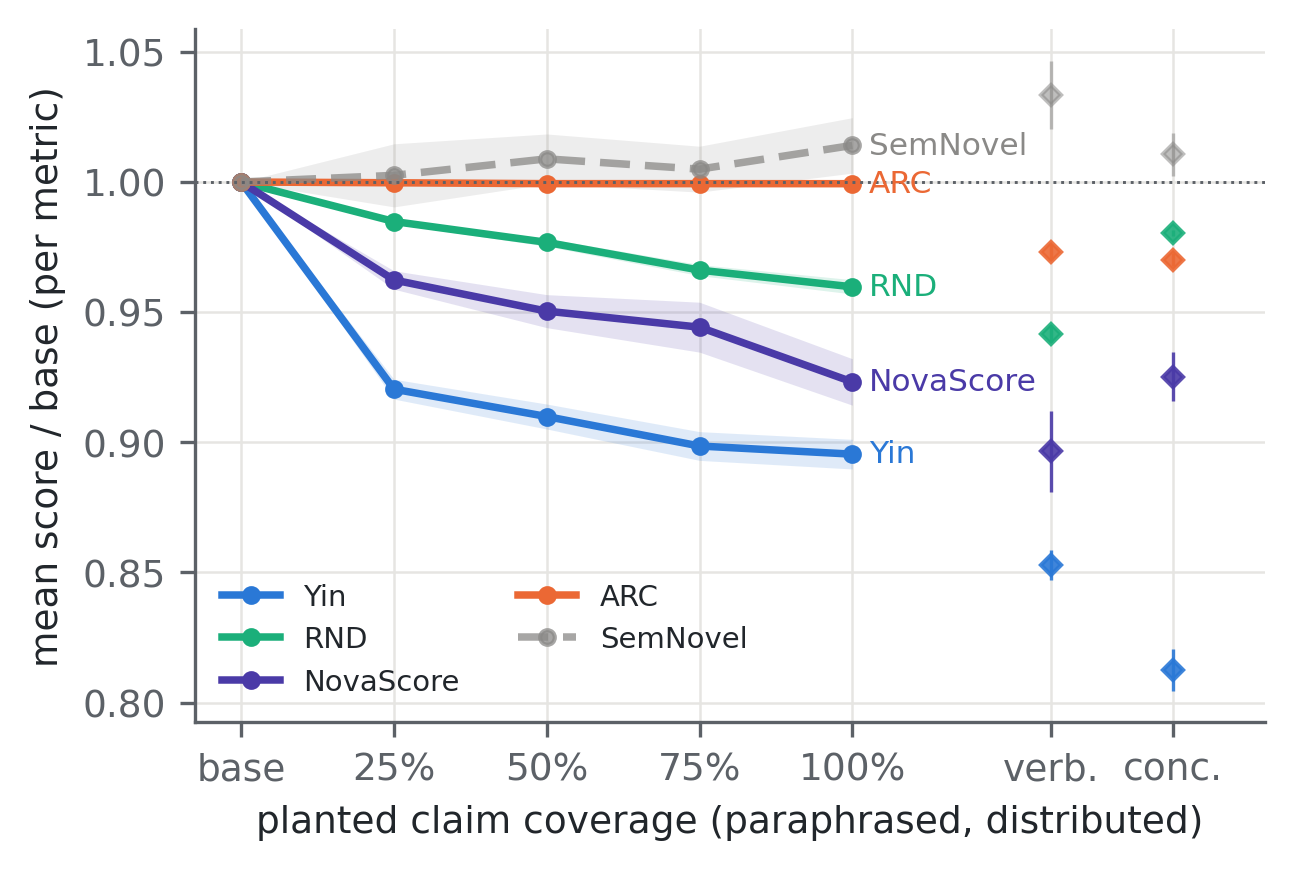}
\caption{R-partial coverage gradient, each system's macro mean score normalized by its own base score (native-scale means in Table~\ref{tab:rp-gradient}). verb.\ = verbatim contrast, conc.\ = concentrated contrast; bands and bars show $\pm 1$ SEM across the ten tasks. SemNovel's upward drift is an artifact of its per-pool t-SNE refit (Appendix~\ref{app:attribution}).}
\label{fig:rp-gradient}
\end{figure}

Although the mean scores of Yin, RND, and NovaScore all decline monotonically as coverage rises (see Figure~\ref{fig:rp-gradient}), the contrast probes help us separate the mechanisms responsible. Yin drops further under verbatim planting than under full paraphrased coverage, and further still when the claims are concentrated in one paper, pointing to surface and single-document matching. NovaScore, on the other hand, has the same movement whether the claims are concentrated or distributed, responding to the coverage itself. However, it detects exact copies at only 0.73, showing the claim-level and document-level designs fail in complementary ways. Additionally, planting a \textit{different} paper's claims into the same host papers leaves Yin and RND with the same scores (mean score 1.00 of base across all ten tasks), indicating the R-partial declines measure coverage of the focal paper, and not generic perturbation of its neighbors. These contrasts in behavior reveal the mechanisms behind the failures, making them more actionable.

The same is true for the AI for science systems. The AI Scientist judge detects copies and paraphrases about equally (0.70 and 0.73), but it does not beat chance for claim-level coverage. RAG-Novelty fails all redundancy probes including the exact copy probe (0.28), confirming its scoring ignores retrieval. However, while its 1--10 integer output does not move on over half of all papers, its rare movements do lean correct (Table~\ref{tab:ties}), suggesting RAG-Novelty's issue is primarily insensitivity. Since ARC is only a lexical check, it is unsurprising that it can detect exact copies, and partly paraphrases (0.60), but fails everything finer. Its remaining failures are ties rather than wrong-way movements (Table~\ref{tab:ties}).

\paragraph{Si tournament (\S\ref{sec:systems}).}
For the Si protocol, which we probe through rank displacement, we find that the focal paper's rank drops when a copy of it is inserted, but that the rank drops similarly for both a paraphrase and for an unrelated paper (Table~\ref{tab:si-control}). We conclude that the pairwise tournament, which ranks by predicted acceptance rather than explicit novelty (Appendix~\ref{app:adapters}), reacts to \textit{any} insertion and is not sensitive to redundancy.

\paragraph{OpenNovelty attribution.}
OpenNovelty behaves differently. On 50 focal papers each from \texttt{drugdiscovery} and \texttt{codegeneration}, it correctly attributes 100\% of exact duplicates, 96--100\% of paraphrases, and 76--78\% of distributed claim plants, with zero false fires on 99 unmanipulated controls. The fraction of contributions its verdicts mark as refuted rises monotonically in both tasks, from 0.12 on controls to 0.98--0.99 under exact duplicates. We scoped this evaluation to two tasks for cost (see Appendix~\ref{app:ledger}).

Constructively, we see that structured use of LLMs (as with NovaScore or OpenNovelty) can achieve claim-level sensitivity, but embedding and lexical approaches are still strong, cheap detectors of exact or paraphrased unoriginality. The rare four-step gradient ($\leq 0.08$ everywhere) mostly reflects the low chance floor of chained checks.

\subsubsection{Redundancy in the Wild}
\label{sec:inthewild}

To show that our axiom of redundancy applies in the wild, we again draw upon ICLR review data, this time the review texts rather than the scores. From reviewer comments, we mine 197 unique $(A, Y)$ pairs where a reviewer explicitly disputes the novelty of work $A$ based on prior work $Y$ (Appendix~\ref{app:inthewild}). For the relevant systems, we score these pairs, evaluating $s(A,\, C \cup \{Y\}) < s(A,\, C \setminus \{Y\})$. Because a one-paper pool insertion would only be observable for continuous scorers that don't refit representations per pool, the systems in question are Yin, RND, and NovaScore.

As our probes would predict, Yin passes only 25\% of pairs, and every failure is an exact tie, because the minimum distance moves only if $Y$ becomes $A$'s single nearest neighbor. RND, having a similar embedding approach, behaves the same way (Table~\ref{tab:inthewild}). However, the signal \textit{is} present, since the reviewer-named $Y$ has median rank 8 among $A$'s pool neighbors; aggregating over the top ${\sim}20$ neighbors would recover it for over half the pairs. For NovaScore, on the other hand, the 24\% pass rate is noise rather than detection. Its non-ties split exactly 47/47 (sign test $p \approx 1$); this claim-level implementation does not yet register a real relevant paper in the pool.

\subsection{Relevance}
\label{sec:p2results}

Our relevance probes are a test of pool sensitivity. The coarsest and easiest probe swaps the entire base pool for a distant task's pool; a system which fails this probe is not meaningfully conditioned on its pool. However, on this test the simple systems generally do well, and two more complex LLM systems do not (with SemNovel and the AI Scientist judge being respective exceptions; see Appendix~\ref{app:attribution} for more on SemNovel). Yin, RND, and the AI Scientist judge respond correctly on every relevance probe, and FastTextLOF fails only the cited-paper removal. We note that when an LLM system receives retrieved papers as text, as the AI Scientist judge does, it can act on them, but retrieval reduced to summary statistics (as in RAG-Novelty, which fails all five, largely by not moving at all) is not useful. For NovaScore, which has a finer comparison unit, the score moves only when pool content is close enough to entail the focal paper's claims, and there are ties on 36\% of papers as a result.

The performance of the various systems degrades as the relevance difference becomes finer. On the far-versus-near probe, for instance, RND's bounded percentiles saturate; 80\% of its failures are exact ties at its ceiling of 100, and it scores a mere 0.59 where Yin's raw distances score 0.90. Additionally, every strong system loses ground on the citation probes, since removing a paper's cited prior art changes only a small, nearby part of the pool. The deployment task for these novelty systems \textit{is} judging a paper against actual predecessors, which is exactly where they currently falter.

\subsection{Temporality}

The performance of novelty systems on the axiom of temporality is noticeably weak compared to the other two. Many of the systems score slightly above chance in separating adjacent pairs of windows, but the performance shrinks as we approach the present; this is to some degree expected, because equal-count windows near the present span only months of a fast-growing literature. The monotone probe requirement is naturally far stricter, and no system exceeds 0.24. However, the aggregate trend is real; Page tests find this decline significant on 10/10 tasks for Yin and 9/10 for RND (Table~\ref{tab:page}).

Subsumption is the hardest probe in the set, as \textit{no} novelty system reliably scores a paper lower against the future slice than against the newest past window. Scoring the future slice against a \textit{random base pool sample of the same size}, Yin passes 50\% of papers and RND 28\%, so RND's apparent pass against the full base pool (0.57 in Table~\ref{tab:main}) is a pool-size artifact. We also validate the probe's premise. We identify focal papers whose future slice \textit{contains work citing them} and find that for this 22\% of papers, every system's pass rate against the newest window remains at or below chance (Appendix~\ref{app:premises}).

Taken together with the window results, this suggests that what the systems weakly detect is the \textit{era} of a pool rather than its content. Accumulated content should make the focal paper look less novel against later literature, while linguistic distance grows in both temporal directions. However, Yin, the strongest trend detector, behaves symmetrically instead, scoring the focal paper higher against the post-focal slice than against the newest past window (0.259 vs 0.252 on average, higher than base for 75\% of papers), even though that slice contains the literature most likely to absorb the paper.

\section{Analysis and Discussion}
\label{sec:satisfiability}

No system does well across all of our axioms and probes, and even Yin, clearly the strongest system in Table~\ref{tab:main}, passes many only modestly above chance. The failures are diagnostic, however, of the kinds of novelty each architecture captures, and the individual capabilities largely exist in separate architectures. NovaScore compares claim by claim, so it is the only system whose score does not depend on whether the planted claims sit in one pool paper or many. This property, however, inadequately addresses the real-world understanding of scientific novelty, in which assembling known claims into a coherent whole is itself a contribution. The embedding metrics remain cheap, reliable detectors of copies, paraphrases, and pool relevance. Future novelty components should take these complementary strengths into account.

The notable exception in performance is the temporality subsumption probe, which remains at best weakly captured by the systems we evaluate. Novelty, of course, is time-indexed, and an idea that is novel at some point in time may be scooped before it is executed; we may imagine this is an area where LLM judges, having parametric knowledge, could excel given the correct scaffolding. More generally, a system's results on the benchmark form a \textit{behavioral profile} of the system's successes and failures. They show clearly where a system can improve, how it might be combined effectively with others, and how future systems may be better designed to address current gaps.

\section{Conclusion}

We introduced an axiomatic benchmark for scientific novelty metrics as an objective, detailed, and label-free way to compare them, testing whether scores move correctly under controlled changes to the reference pool. It complements validation against human ratings, whose most direct signal we show to be inseparable from quality, and its per-probe results form an interpretable profile of each metric. The results run against the natural expectation that LLM-based systems have made simpler metrics obsolete. Embedding and lexical metrics can cheaply and robustly detect copies, paraphrases, and changes in pool relevance, often beating the LLM systems. Deliberate, structured LLM usage, however, can check a paper claim by claim against the pool, and LLMs' parametric knowledge may yet address subsumption, the hardest probe. Our setting is also deliberately easy, asking only for the \textit{direction} of a change against a pool of modest size; a system that fails here will fail much harder in deployment, and much of novelty evaluation remains open. We release the benchmark, pools, and code.

\section*{Limitations}
\label{sec:limitations}
Our construction involves several proxies. For instance, the pools are task-scoped and contain only one to two thousand papers. This makes it impossible to test how systems would perform against the real scientific literature, which is much larger. Similarly, field distance is only a proxy for low relevance (although we attempt to validate it in this instance). A better proxy may need to be constructed when investigating cases where there is, say, a lot of interdisciplinary transfer involved. Finally, all of our tasks are AI research tasks. This is an inherent limit of the tagged corpus we draw from.

For LLM systems, our sandboxing controls what a judge retrieves, but not what it already knows. Our probes cannot remove the parametric knowledge of the LLM models; they can only measure sensitivity to the pool \textit{on top of} this knowledge. Additionally, some systems are modified quite heavily so that we can evaluate fairly. For instance, the fixed-frame control pass elaborated on in Appendix~\ref{app:attribution} is only strictly relevant for the modified variant we constructed rather than for the deployed method. Finally, the axioms are \textit{necessary} conditions. A system that passes every probe can still be a poor judge of novelty, and we make no claim to the contrary.

\section*{Ethics Statement}

We want to encourage the development of improved novelty metrics to assist rather than replace human judgment. However, we recognize that all such metrics can be gamed, although one of the aims of this work is to make such gaming more challenging and to better align metrics with the current human understanding of scientific novelty. Nevertheless, any kind of automation of novelty screening has some ethical considerations, and it is possible that these metrics may be biased, so they should be used carefully.

\section*{Acknowledgments}

We thank Caleb Biddulph for comments and discussion on an earlier draft, Eric Modesitt for thoughts on the organization of our axiomatic framework, and other members of the TIMAN Lab at UIUC for their feedback. Miri Liu is supported by the Amazon AI PhD Fellowship. This work is supported in part by the National Science Foundation (NSF) under Grants 2229612 and 2433308. Any opinions, findings, and conclusions or recommendations expressed in this material are those of the authors and do not necessarily reflect the views of the National Science Foundation.

\bibliography{custom}

@inproceedings{novascore,
    title = "{N}ov{AS}core: A New Automated Metric for Evaluating Document Level Novelty",
    author = "Ai, Lin  and
      Gong, Ziwei  and
      Deshpande, Harshsaiprasad  and
      Johnson, Alexander  and
      Phung, Emmy  and
      Emami, Ahmad  and
      Hirschberg, Julia",
    editor = "Rambow, Owen  and
      Wanner, Leo  and
      Apidianaki, Marianna  and
      Al-Khalifa, Hend  and
      Eugenio, Barbara Di  and
      Schockaert, Steven",
    booktitle = "Proceedings of the 31st International Conference on Computational Linguistics",
    month = jan,
    year = "2025",
    address = "Abu Dhabi, UAE",
    publisher = "Association for Computational Linguistics",
    url = "https://aclanthology.org/2025.coling-main.234/",
    pages = "3479--3494"
}

@article{fasttext,
  author   = {Daeseong Jeon and Junyoup Lee and Joon Mo Ahn and Changyong Lee},
  title    = {Measuring the novelty of scientific publications: A fastText and local outlier factor approach},
  journal  = {Journal of Informetrics},
  volume   = {17},
  number   = {4},
  pages    = {101450},
  year     = {2023},
  month = nov,
  doi      = {10.1016/j.joi.2023.101450},
  url      = {https://doi.org/10.1016/j.joi.2023.101450}
}

@misc{rnd,
      title={Enabling AI Scientists to Recognize Innovation: A Domain-Agnostic Algorithm for Assessing Novelty}, 
      author={Yao Wang and Mingxuan Cui and Arthur Jiang and Jun Yan},
      year={2025},
      eprint={2503.01508},
      archivePrefix={arXiv},
      primaryClass={cs.AI},
      url={https://arxiv.org/abs/2503.01508}, 
}

@article{yin,
  author  = {Yin, D. and Wu, Z. and Yokota, K. and Matsumoto, K. and Shibayama, S.},
  title   = {Identify novel elements of knowledge with word embedding},
  journal = {PLOS ONE},
  volume  = {18},
  number  = {6},
  pages   = {e0284567},
  year    = {2023},
  doi     = {10.1371/journal.pone.0284567},
  url     = {https://doi.org/10.1371/journal.pone.0284567}
}

@article{shibayama2021novelty,
  title     = {Measuring novelty in science with word embedding},
  author    = {Shibayama, Sotaro and Yin, Deyun and Matsumoto, Kuniko},
  journal   = {PLOS ONE},
  volume    = {16},
  number    = {7},
  pages     = {e0254034},
  year      = {2021},
  month     = {Jul},
  doi       = {10.1371/journal.pone.0254034},
  pmid      = {34214135},
  pmcid     = {PMC8253414},
  url={https://doi.org/10.1371/journal.pone.0254034}
}

@article{semnovel,
  author  = {Peng, Xueqing and Xie, Yutong and He, Huan and Ondov, Brian and Raja, Kalpana and Liu, Qijia and Mei, Qiaozhu and Xu, Hua},
  title   = {SemNovel - A new approach to detecting semantic novelty of biomedical publications using embeddings of large language models},
  journal = {Journal of Biomedical Informatics},
  volume  = {172},
  pages   = {104952},
  year    = {2025},
  month   = dec,
  doi     = {10.1016/j.jbi.2025.104952},
  url     = {https://doi.org/10.1016/j.jbi.2025.104952},
  pmid    = {41242670},
}

@misc{yamada2025aiscientistv2workshoplevelautomated,
      title={The AI Scientist-v2: Workshop-Level Automated Scientific Discovery via Agentic Tree Search}, 
      author={Yutaro Yamada and Robert Tjarko Lange and Cong Lu and Shengran Hu and Chris Lu and Jakob Foerster and Jeff Clune and David Ha},
      year={2025},
      eprint={2504.08066},
      archivePrefix={arXiv},
      primaryClass={cs.AI},
      url={https://arxiv.org/abs/2504.08066}, 
}

@misc{moussa2026scholarevalresearchideaevaluation,
      title={ScholarEval: Research Idea Evaluation Grounded in Literature}, 
      author={Hanane Nour Moussa and Patrick Queiroz Da Silva and Daniel Adu-Ampratwum and Alyson East and Zitong Lu and Nikki Puccetti and Mingyi Xue and Huan Sun and Bodhisattwa Prasad Majumder and Sachin Kumar},
      year={2026},
      eprint={2510.16234},
      archivePrefix={arXiv},
      primaryClass={cs.AI},
      url={https://arxiv.org/abs/2510.16234}, 
}

@inproceedings{beel2025evaluatingaiscientist,
  title={Evaluating Sakana's AI Scientist: Bold Claims, Mixed Results, and a Promising Future?},
  author={Beel, Joeran and Kan, Min-Yen and Baumgart, Moritz},
  booktitle={ACM SIGIR Forum},
  volume={59},
  pages={1--20},
  year={2025},
  organization={ACM New York, NY, USA}
}

@misc{jiang2026hindsightevaluatingllmgeneratedresearch,
      title={HindSight: Evaluating LLM-Generated Research Ideas via Future Impact}, 
      author={Bo Jiang},
      year={2026},
      eprint={2603.15164},
      archivePrefix={arXiv},
      primaryClass={cs.CL},
      url={https://arxiv.org/abs/2603.15164}, 
}

@article{chen1998metamorphic,
  title={Metamorphic testing: a new approach for generating next test cases. Department of Computer Science, Hong Kong University of Science and Technology},
  author={Chen, Tsong Yueh and Cheung, Shing Chi and Yiu, Shiu Ming},
  journal={Tech. Rep. HKUST-CS98--01, Tech. Rep.},
  year={1998}
}

@misc{sinhahajari2026limitsllmasjudgescientificnovelty,
      title={On the Limits of LLM-as-Judge for Scientific Novelty Assessment}, 
      author={Soumitra Sinhahajari and Navonil Majumder and Soujanya Poria},
      year={2026},
      eprint={2606.12071},
      archivePrefix={arXiv},
      primaryClass={cs.DL},
      url={https://arxiv.org/abs/2606.12071}, 
}

@misc{schopf2026ideanovelautomatedbenchmark,
      title={Is this Idea Novel? An Automated Benchmark for Judgment of Research Ideas}, 
      author={Tim Schopf and Michael Färber},
      year={2026},
      eprint={2603.10303},
      archivePrefix={arXiv},
      primaryClass={cs.CL},
      url={https://arxiv.org/abs/2603.10303}, 
}

@misc{wu2026novbenchevaluatinglargelanguage,
      title={NovBench: Evaluating Large Language Models on Academic Paper Novelty Assessment}, 
      author={Wenqing Wu and Yi Zhao and Yuzhuo Wang and Siyou Li and Juexi Shao and Yunfei Long and Chengzhi Zhang},
      year={2026},
      eprint={2604.11543},
      archivePrefix={arXiv},
      primaryClass={cs.CL},
      url={https://arxiv.org/abs/2604.11543}, 
}

@misc{zhang2026opennoveltyllmpoweredagenticverifiable,
      title={OpenNovelty: An LLM-powered Agentic System for Verifiable Scholarly Novelty Assessment}, 
      author={Ming Zhang and Kexin Tan and Yueyuan Huang and Yujiong Shen and Chunchun Ma and Li Ju and Xinran Zhang and Yuhui Wang and Wenqing Jing and Jingyi Deng and Huayu Sha and Binze Hu and Jingqi Tong and Changhao Jiang and Yage Geng and Yuankai Ying and Yue Zhang and Zhangyue Yin and Zhiheng Xi and Shihan Dou and Tao Gui and Qi Zhang and Xuanjing Huang},
      year={2026},
      eprint={2601.01576},
      archivePrefix={arXiv},
      primaryClass={cs.IR},
      url={https://arxiv.org/abs/2601.01576}, 
}

@misc{si2024llmsgeneratenovelresearch,
      title={Can LLMs Generate Novel Research Ideas? A Large-Scale Human Study with 100+ NLP Researchers}, 
      author={Chenglei Si and Diyi Yang and Tatsunori Hashimoto},
      year={2024},
      eprint={2409.04109},
      archivePrefix={arXiv},
      primaryClass={cs.CL},
      url={https://arxiv.org/abs/2409.04109}, 
}

@misc{lu2024aiscientistfullyautomated,
      title={The AI Scientist: Towards Fully Automated Open-Ended Scientific Discovery}, 
      author={Chris Lu and Cong Lu and Robert Tjarko Lange and Jakob Foerster and Jeff Clune and David Ha},
      year={2024},
      eprint={2408.06292},
      archivePrefix={arXiv},
      primaryClass={cs.AI},
      url={https://arxiv.org/abs/2408.06292}, 
}

@misc{baek2025researchagentiterativeresearchidea,
      title={ResearchAgent: Iterative Research Idea Generation over Scientific Literature with Large Language Models}, 
      author={Jinheon Baek and Sujay Kumar Jauhar and Silviu Cucerzan and Sung Ju Hwang},
      year={2025},
      eprint={2404.07738},
      archivePrefix={arXiv},
      primaryClass={cs.CL},
      url={https://arxiv.org/abs/2404.07738}, 
}

@misc{keya2025sciideacontextawarescientificideation,
      title={SCI-IDEA: Context-Aware Scientific Ideation Using Token and Sentence Embeddings}, 
      author={Farhana Keya and Gollam Rabby and Prasenjit Mitra and Sahar Vahdati and Sören Auer and Yaser Jaradeh},
      year={2025},
      eprint={2503.19257},
      archivePrefix={arXiv},
      primaryClass={cs.CL},
      url={https://arxiv.org/abs/2503.19257}, 
}

@misc{schmidgall2025agentlaboratoryusingllm,
      title={Agent Laboratory: Using LLM Agents as Research Assistants}, 
      author={Samuel Schmidgall and Yusheng Su and Ze Wang and Ximeng Sun and Jialian Wu and Xiaodong Yu and Jiang Liu and Michael Moor and Zicheng Liu and Emad Barsoum},
      year={2025},
      eprint={2501.04227},
      archivePrefix={arXiv},
      primaryClass={cs.HC},
      url={https://arxiv.org/abs/2501.04227}, 
}

@misc{si2025ideationexecutiongapexecutionoutcomes,
      title={The Ideation-Execution Gap: Execution Outcomes of LLM-Generated versus Human Research Ideas}, 
      author={Chenglei Si and Tatsunori Hashimoto and Diyi Yang},
      year={2025},
      eprint={2506.20803},
      archivePrefix={arXiv},
      primaryClass={cs.CL},
      url={https://arxiv.org/abs/2506.20803}, 
}

@article{luo2022combination,
  title     = {Combination of research questions and methods: A new measurement of scientific novelty},
  author    = {Luo, Zhuoran and Lu, Wei and He, Jiangen and Wang, Yuqi},
  journal   = {Journal of Informetrics},
  volume    = {16},
  number    = {2},
  pages     = {101282},
  year      = {2022},
  month = may,
  doi       = {10.1016/j.joi.2022.101282},
  url       = {https://doi.org/10.1016/j.joi.2022.101282},
  publisher = {Elsevier}
}

@inproceedings{fang_informationretrieval,
author = {Fang, Hui and Tao, Tao and Zhai, ChengXiang},
title = {A formal study of information retrieval heuristics},
year = {2004},
isbn = {1581138814},
publisher = {Association for Computing Machinery},
address = {New York, NY, USA},
url = {https://doi.org/10.1145/1008992.1009004},
doi = {10.1145/1008992.1009004},
booktitle = {Proceedings of the 27th Annual International ACM SIGIR Conference on Research and Development in Information Retrieval},
pages = {49–56},
numpages = {8},
location = {Sheffield, United Kingdom},
series = {SIGIR '04}
}

@article{wu2019large,
  title={Large teams develop and small teams disrupt science and technology},
  author={Wu, Lingfei and Wang, Dashun and Evans, James A},
  journal={Nature},
  volume={566},
  pages={378--382},
  year={2019},
  doi={10.1038/s41586-019-0941-9},
  url={https://doi.org/10.1038/s41586-019-0941-9}
}

@article{Kinney2023TheSS,
  title={The Semantic Scholar Open Data Platform},
  author={Rodney Michael Kinney and Chloe Anastasiades and Russell Authur and Iz Beltagy and Jonathan Bragg and Alexandra Buraczynski and Isabel Cachola and Stefan Candra and Yoganand Chandrasekhar and Arman Cohan and Miles Crawford and Doug Downey and Jason Dunkelberger and Oren Etzioni and Rob Evans and Sergey Feldman and Joseph Gorney and David W. Graham and F.Q. Hu and Regan Huff and Daniel King and Sebastian Kohlmeier and Bailey Kuehl and Michael Langan and Daniel Lin and Haokun Liu and Kyle Lo and Jaron Lochner and Kelsey MacMillan and Tyler C. Murray and Christopher Newell and Smita Rao and Shaurya Rohatgi and Paul Sayre and Shannon Zejiang Shen and Amanpreet Singh and Luca Soldaini and Shivashankar Subramanian and A. Tanaka and Alex D Wade and Linda M. Wagner and Lucy Lu Wang and Christopher Wilhelm and Caroline Wu and Jiangjiang Yang and Angele Zamarron and Madeleine van Zuylen and Daniel S. Weld},
  journal={ArXiv},
  year={2023},
  volume={abs/2301.10140},
  url={https://api.semanticscholar.org/CorpusID:256194545}
}

@inproceedings{busin2013,
author = {Busin, Luca and Mizzaro, Stefano},
title = {Axiometrics: An Axiomatic Approach to Information Retrieval Effectiveness Metrics},
year = {2013},
isbn = {9781450321075},
publisher = {Association for Computing Machinery},
address = {New York, NY, USA},
url = {https://doi.org/10.1145/2499178.2499182},
doi = {10.1145/2499178.2499182},
booktitle = {Proceedings of the 2013 Conference on the Theory of Information Retrieval},
pages = {22–29},
numpages = {8},
location = {Copenhagen, Denmark},
series = {ICTIR '13}
}

@article{Hanson_2024,
   title={The strain on scientific publishing},
   volume={5},
   url={http://dx.doi.org/10.1162/qss_a_00327},
   DOI={10.1162/qss_a_00327},
   number={4},
   journal={Quantitative Science Studies},
   publisher={MIT Press},
   author={Hanson, Mark A. and Barreiro, Pablo Gómez and Crosetto, Paolo and Brockington, Dan},
   year={2024},
   pages={823–843} }

@misc{bornmann2021growthratesmodernscience,
      title={Growth rates of modern science: A latent piecewise growth curve approach to model publication numbers from established and new literature databases}, 
      author={Lutz Bornmann and Robin Haunschild and Ruediger Mutz},
      year={2021},
      eprint={2012.07675},
      archivePrefix={arXiv},
      primaryClass={cs.DL},
      url={https://arxiv.org/abs/2012.07675}, 
}

@misc{tahamtan2018creativitysciencelinkcited,
      title={Creativity in Science and the Link to Cited References: Is the Creative Potential of Papers Reflected in their Cited References?}, 
      author={Iman Tahamtan and Lutz Bornmann},
      year={2018},
      eprint={1806.00224},
      archivePrefix={arXiv},
      primaryClass={cs.DL},
      url={https://arxiv.org/abs/1806.00224}, 
}

@article{zhao2025novelty,
  author    = {Zhao, Y. and Zhang, C.},
  title     = {A review on the novelty measurements of academic papers},
  journal   = {Scientometrics},
  volume    = {130},
  pages     = {727--753},
  year      = {2025},
  month     = {February},
  doi       = {10.1007/s11192-025-05234-0},
  url       = {https://doi.org/10.1007/s11192-025-05234-0},
  received  = {2024-02-23},
  accepted  = {2025-01-07},
  published = {2025-01-27}
}

@article{fontana2020new,
  title={New and atypical combinations: An assessment of novelty and interdisciplinarity},
  author={Fontana, Magda and Iori, Martina and Montobbio, Fabio and Sinatra, Roberta},
  journal={Research Policy},
  volume={49},
  number={7},
  articleno={104063},
  year={2020},
  publisher={Elsevier},
  doi={10.1016/j.respol.2020.104063},
  url={https://doi.org/10.1016/j.respol.2020.104063}
}

@misc{gates2025increasingfragmentationglobalscience,
      title={The increasing fragmentation of global science limits the diffusion of ideas}, 
      author={Alexander J. Gates and Jianjian Gao and Indraneel Mane},
      year={2025},
      eprint={2404.05861},
      archivePrefix={arXiv},
      primaryClass={cs.SI},
      url={https://arxiv.org/abs/2404.05861}, 
}

@article{evans2008,
author = {James A. Evans },
title = {Electronic Publication and the Narrowing of Science and Scholarship},
journal = {Science},
volume = {321},
number = {5887},
pages = {395-399},
year = {2008},
doi = {10.1126/science.1150473},
URL = {https://www.science.org/doi/abs/10.1126/science.1150473},
eprint = {https://www.science.org/doi/pdf/10.1126/science.1150473}}

@article{park2023papers,
  author    = {Park, Michael and Leahey, Erin and Funk, Russell J.},
  title     = {Papers and patents are becoming less disruptive over time},
  journal   = {Nature},
  volume    = {613},
  pages     = {138--144},
  year      = {2023},
  month     = jan,
  doi       = {10.1038/s41586-022-05543-x},
  url       = {https://doi.org/10.1038/s41586-022-05543-x},
  received  = {2022-02-14},
  accepted  = {2022-11-08},
  published = {2023-01-04}
}

@misc{spiliopoulou2025playfavoritesstatisticalmethod,
      title={Play Favorites: A Statistical Method to Measure Self-Bias in LLM-as-a-Judge}, 
      author={Evangelia Spiliopoulou and Riccardo Fogliato and Hanna Burnsky and Tamer Soliman and Jie Ma and Graham Horwood and Miguel Ballesteros},
      year={2025},
      eprint={2508.06709},
      archivePrefix={arXiv},
      primaryClass={cs.CL},
      url={https://arxiv.org/abs/2508.06709}, 
}

@inproceedings{shi-etal-2025-judging,
    title = "Judging the Judges: A Systematic Study of Position Bias in {LLM}-as-a-Judge",
    author = "Shi, Lin  and
      Ma, Chiyu  and
      Liang, Wenhua  and
      Diao, Xingjian  and
      Ma, Weicheng  and
      Vosoughi, Soroush",
    editor = "Inui, Kentaro  and
      Sakti, Sakriani  and
      Wang, Haofen  and
      Wong, Derek F.  and
      Bhattacharyya, Pushpak  and
      Banerjee, Biplab  and
      Ekbal, Asif  and
      Chakraborty, Tanmoy  and
      Singh, Dhirendra Pratap",
    booktitle = "Proceedings of the 14th International Joint Conference on Natural Language Processing and the 4th Conference of the Asia-Pacific Chapter of the Association for Computational Linguistics",
    month = dec,
    year = "2025",
    address = "Mumbai, India",
    publisher = "The Asian Federation of Natural Language Processing and The Association for Computational Linguistics",
    url = "https://aclanthology.org/2025.ijcnlp-long.18/",
    doi = "10.18653/v1/2025.ijcnlp-long.18",
    pages = "292--314",
    ISBN = "979-8-89176-298-5"
}

@misc{bornmann2019novelty,
      title={Do we measure novelty when we analyze unusual combinations of cited references? A validation study of bibliometric novelty indicators based on F1000Prime data}, 
      author={Lutz Bornmann and Alexander Tekles and Helena H. Zhang and Fred Y. Ye},
      year={2019},
      eprint={1910.03233},
      archivePrefix={arXiv},
      primaryClass={cs.DL},
      url={https://arxiv.org/abs/1910.03233}, 
}

@article{mishra2016novelty,
  title     = {Quantifying Conceptual Novelty in the Biomedical Literature},
  author    = {Mishra, Shubhanshu and Torvik, Vetle I.},
  journal   = {D-Lib magazine : the magazine of the Digital Library Forum},
  volume    = {22},
  number    = {9-10},
  year      = {2016},
  month     = {Sep-Oct},
  doi       = {10.1045/september2016-mishra},
  pmid      = {27942200},
  pmcid     = {PMC5142764},
  url={https://doi.org/10.1045/september2016-mishra}
}

@inproceedings{sendhilkumar2013novelty,
  title={Novelty Detection via Topic Modeling in Research Articles},
  author={S. Sendhilkumar and Nachiyar S Nandhini and Mahalakshmi G.S},
  year={2013},
  booktitle={Third International Conference on Computer Science, Engineering \& Applications},
  url={https://api.semanticscholar.org/CorpusID:2399905}
}

@article{ghosal2022novelty,
    title = "Novelty Detection: A Perspective from Natural Language Processing",
    author = "Ghosal, Tirthankar  and
      Saikh, Tanik  and
      Biswas, Tameesh  and
      Ekbal, Asif  and
      Bhattacharyya, Pushpak",
    journal = "Computational Linguistics",
    volume = "48",
    number = "1",
    month = mar,
    year = "2022",
    address = "Cambridge, MA",
    publisher = "MIT Press",
    url = "https://aclanthology.org/2022.cl-1.3/",
    doi = "10.1162/coli_a_00429",
    pages = "77--117"
}

@inproceedings{ghosal-etal-2018-novelty,
    title = "Novelty Goes Deep. A Deep Neural Solution To Document Level Novelty Detection",
    author = "Ghosal, Tirthankar  and
      Edithal, Vignesh  and
      Ekbal, Asif  and
      Bhattacharyya, Pushpak  and
      Tsatsaronis, George  and
      Chivukula, Srinivasa Satya Sameer Kumar",
    editor = "Bender, Emily M.  and
      Derczynski, Leon  and
      Isabelle, Pierre",
    booktitle = "Proceedings of the 27th International Conference on Computational Linguistics",
    month = aug,
    year = "2018",
    address = "Santa Fe, New Mexico, USA",
    publisher = "Association for Computational Linguistics",
    url = "https://aclanthology.org/C18-1237/",
    pages = "2802--2813"
}

@inproceedings{amigo2013,
author = {Amig\'{o}, Enrique and Gonzalo, Julio and Verdejo, Felisa},
title = {A general evaluation measure for document organization tasks},
year = {2013},
isbn = {9781450320344},
publisher = {Association for Computing Machinery},
address = {New York, NY, USA},
url = {https://doi.org/10.1145/2484028.2484081},
doi = {10.1145/2484028.2484081},
booktitle = {Proceedings of the 36th International ACM SIGIR Conference on Research and Development in Information Retrieval},
pages = {643–652},
numpages = {10},
location = {Dublin, Ireland},
series = {SIGIR '13}
}

@inproceedings{sebastiani2015,
author = {Sebastiani, Fabrizio},
title = {An Axiomatically Derived Measure for the Evaluation of Classification Algorithms},
year = {2015},
isbn = {9781450338332},
publisher = {Association for Computing Machinery},
address = {New York, NY, USA},
url = {https://doi.org/10.1145/2808194.2809449},
doi = {10.1145/2808194.2809449},
booktitle = {Proceedings of the 2015 International Conference on The Theory of Information Retrieval},
pages = {11–20},
numpages = {10},
location = {Northampton, Massachusetts, USA},
series = {ICTIR '15}
}

@inproceedings{amplayo2019evaluating,
    title = "Evaluating Research Novelty Detection: Counterfactual Approaches",
    author = "Amplayo, Reinald Kim  and
      Hwang, Seung-won  and
      Song, Min",
    editor = "Ustalov, Dmitry  and
      Somasundaran, Swapna  and
      Jansen, Peter  and
      Glava{\v{s}}, Goran  and
      Riedl, Martin  and
      Surdeanu, Mihai  and
      Vazirgiannis, Michalis",
    booktitle = "Proceedings of the Thirteenth Workshop on Graph-Based Methods for Natural Language Processing (TextGraphs-13)",
    month = nov,
    year = "2019",
    address = "Hong Kong",
    publisher = "Association for Computational Linguistics",
    url = "https://aclanthology.org/D19-5315/",
    doi = "10.18653/v1/D19-5315",
    pages = "124--133"
}

@misc{yang2024largelanguagemodelsautomated,
      title={Large Language Models for Automated Open-domain Scientific Hypotheses Discovery}, 
      author={Zonglin Yang and Xinya Du and Junxian Li and Jie Zheng and Soujanya Poria and Erik Cambria},
      year={2024},
      eprint={2309.02726},
      archivePrefix={arXiv},
      primaryClass={cs.CL},
      url={https://arxiv.org/abs/2309.02726}, 
}

@inproceedings{soboroff-harman-2005-novelty,
    title = "Novelty Detection: The {TREC} Experience",
    author = "Soboroff, Ian  and
      Harman, Donna",
    editor = "Mooney, Raymond  and
      Brew, Chris  and
      Chien, Lee-Feng  and
      Kirchhoff, Katrin",
    booktitle = "Proceedings of Human Language Technology Conference and Conference on Empirical Methods in Natural Language Processing",
    month = oct,
    year = "2005",
    address = "Vancouver, British Columbia, Canada",
    publisher = "Association for Computational Linguistics",
    url = "https://aclanthology.org/H05-1014/",
    pages = "105--112"
}

@Article{wang2024effective,
journal={Journal of Informetrics},
author={Wang, Zhongyi and Zhang, Haoxuan and Chen, Jiangping and Chen, Haihua},
title={An effective framework for measuring the novelty of scientific articles through integrated topic modeling and cloud model},
year={2024},
volume={18},
number={4},
doi={10.1016/j.joi.2024.101587},
url={https://ideas.repec.org/a/eee/infome/v18y2024i4s1751157724000993.html},
}

@article{ruan2025,
author = {Xuanmin Ruan and Weiyi Ao and Dongqing Lyu and Ying Cheng and Jiang Li},
title ={Effect of the topic-combination novelty on the disruption and impact of scientific articles: Evidence from PubMed},

journal = {Journal of Information Science},
volume = {51},
number = {5},
pages = {1033-1046},
year = {2025},
doi = {10.1177/01655515231161133},

URL = { 
    
        https://doi.org/10.1177/01655515231161133
    
    

},
eprint = { 
    
        https://doi.org/10.1177/01655515231161133
    
    

}
}

@article{
uzzi2013,
author = {Brian Uzzi  and Satyam Mukherjee  and Michael Stringer  and Ben Jones },
title = {Atypical Combinations and Scientific Impact},
journal = {Science},
volume = {342},
number = {6157},
pages = {468-472},
year = {2013},
doi = {10.1126/science.1240474},
URL = {https://www.science.org/doi/abs/10.1126/science.1240474},
eprint = {https://www.science.org/doi/pdf/10.1126/science.1240474}}

@misc{weng2025deepscientistadvancingfrontierpushingscientific,
      title={DeepScientist: Advancing Frontier-Pushing Scientific Findings Progressively}, 
      author={Yixuan Weng and Minjun Zhu and Qiujie Xie and Qiyao Sun and Zhen Lin and Sifan Liu and Yue Zhang},
      year={2025},
      eprint={2509.26603},
      archivePrefix={arXiv},
      primaryClass={cs.CL},
      url={https://arxiv.org/abs/2509.26603}, 
}

@misc{tang2025airesearcherautonomousscientificinnovation,
      title={AI-Researcher: Autonomous Scientific Innovation}, 
      author={Jiabin Tang and Lianghao Xia and Zhonghang Li and Chao Huang},
      year={2025},
      eprint={2505.18705},
      archivePrefix={arXiv},
      primaryClass={cs.AI},
      url={https://arxiv.org/abs/2505.18705}, 
}

@misc{li2024chainideasrevolutionizingresearch,
      title={Chain of Ideas: Revolutionizing Research Via Novel Idea Development with LLM Agents}, 
      author={Long Li and Weiwen Xu and Jiayan Guo and Ruochen Zhao and Xingxuan Li and Yuqian Yuan and Boqiang Zhang and Yuming Jiang and Yifei Xin and Ronghao Dang and Deli Zhao and Yu Rong and Tian Feng and Lidong Bing},
      year={2024},
      eprint={2410.13185},
      archivePrefix={arXiv},
      primaryClass={cs.AI},
      url={https://arxiv.org/abs/2410.13185}, 
}

@misc{ghareeb2025robinmultiagentautomatingscientific,
      title={Robin: A multi-agent system for automating scientific discovery}, 
      author={Ali Essam Ghareeb and Benjamin Chang and Ludovico Mitchener and Angela Yiu and Caralyn J. Szostkiewicz and Jon M. Laurent and Muhammed T. Razzak and Andrew D. White and Michaela M. Hinks and Samuel G. Rodriques},
      year={2025},
      eprint={2505.13400},
      archivePrefix={arXiv},
      primaryClass={cs.AI},
      url={https://arxiv.org/abs/2505.13400}, 
}

@misc{weng2025cycleresearcherimprovingautomatedresearch,
      title={CycleResearcher: Improving Automated Research via Automated Review}, 
      author={Yixuan Weng and Minjun Zhu and Guangsheng Bao and Hongbo Zhang and Jindong Wang and Yue Zhang and Linyi Yang},
      year={2025},
      eprint={2411.00816},
      archivePrefix={arXiv},
      primaryClass={cs.CL},
      url={https://arxiv.org/abs/2411.00816}, 
}

@misc{qiao2026innoevalresearchideaevaluation,
      title={InnoEval: On Research Idea Evaluation as a Knowledge-Grounded, Multi-Perspective Reasoning Problem}, 
      author={Shuofei Qiao and Yunxiang Wei and Xuehai Wang and Bin Wu and Boyang Xue and Ningyu Zhang and Hossein A. Rahmani and Yanshan Wang and Qiang Zhang and Keyan Ding and Jeff Z. Pan and Huajun Chen and Emine Yilmaz},
      year={2026},
      eprint={2602.14367},
      archivePrefix={arXiv},
      primaryClass={cs.CL},
      url={https://arxiv.org/abs/2602.14367}, 
}

@misc{lin2024evaluatingenhancinglargelanguage,
      title={Evaluating and Enhancing Large Language Models for Novelty Assessment in Scholarly Publications}, 
      author={Ethan Lin and Zhiyuan Peng and Yi Fang},
      year={2024},
      eprint={2409.16605},
      archivePrefix={arXiv},
      primaryClass={cs.CL},
      url={https://arxiv.org/abs/2409.16605}, 
}

@misc{liu2026autoresearchclawselfreinforcingautonomousresearch,
      title={AutoResearchClaw: Self-Reinforcing Autonomous Research with Human-AI Collaboration}, 
      author={Jiaqi Liu and Shi Qiu and Mairui Li and Bingzhou Li and Haonian Ji and Siwei Han and Xinyu Ye and Peng Xia and Zihan Dong and Meng Chen and Congyu Zhang and Letian Zhang and Guiming Chen and Haoqin Tu and Xinyu Yang and Lu Feng and Xujiang Zhao and Haifeng Chen and Jiawei Zhou and Xiao Wang and Weitong Zhang and Hongtu Zhu and Yun Li and Jieru Mei and Hongliang Fei and Jiaheng Zhang and Linjie Li and Linjun Zhang and Yuyin Zhou and Sheng Wang and Caiming Xiong and James Zou and Zeyu Zheng and Cihang Xie and Mingyu Ding and Huaxiu Yao},
      year={2026},
      eprint={2605.20025},
      archivePrefix={arXiv},
      primaryClass={cs.AI},
      url={https://arxiv.org/abs/2605.20025}, 
}

@misc{panickssery2024llmevaluatorsrecognizefavor,
      title={LLM Evaluators Recognize and Favor Their Own Generations}, 
      author={Arjun Panickssery and Samuel R. Bowman and Shi Feng},
      year={2024},
      eprint={2404.13076},
      archivePrefix={arXiv},
      primaryClass={cs.CL},
      url={https://arxiv.org/abs/2404.13076}, 
}

@misc{ye2024justiceprejudicequantifyingbiases,
      title={Justice or Prejudice? Quantifying Biases in LLM-as-a-Judge}, 
      author={Jiayi Ye and Yanbo Wang and Yue Huang and Dongping Chen and Qihui Zhang and Nuno Moniz and Tian Gao and Werner Geyer and Chao Huang and Pin-Yu Chen and Nitesh V Chawla and Xiangliang Zhang},
      year={2024},
      eprint={2410.02736},
      archivePrefix={arXiv},
      primaryClass={cs.CL},
      url={https://arxiv.org/abs/2410.02736}, 
}

\appendix

\section{AI Usage Disclosure}
LLM agents (e.g., Claude Code) were used to assist with experiment and analysis code, to advise on paper structure and clarity, to format results tables, and to generate the released code README. All scientific contributions, including the axiom and probe formulation, experimental design, and data collection, were devised by the authors.

\section{Motivating Study Details}
\label{app:motivating}

We balance samples across the two years, with consensus defined as reviewer standard deviation $\leq 0.5$ on the chosen field. Each paper's reference pool is the union of its resolved references and Semantic Scholar bulk-search results for three LLM-extracted keywords, time-filtered to before the paper's submission.

All systems score only the title and abstract. We compute Spearman's $\rho$ against the reviewer field mean and ROC AUC against the high/low bucket. The two samples overlap (because many high technical novelty papers are also high empirical novelty papers and vice versa for low novelty) and deduplicate to n=344 unique papers. The median reference pool holds roughly 320 papers. We do not consider the Si pairwise tournament and OpenNovelty because they do not output a per-paper score.

In the technical novelty sample, 88 of the 100 high-novelty papers were accepted and 99 of the 100 low-novelty papers were rejected. For the empirical novelty sample, the split is 95/5 and 1/99 instead. \textit{Only} RAG-Novelty's score predicts the decision itself. Its acceptance AUC is 0.83, against 0.54 for NovaScore, 0.55 for the AI Scientist judge, and 0.45 for ARC.

For the RAG-Novelty shuffle ablation, every paper is rescored with its retrieval summary swapped for that of another paper. The correlations are unchanged (technical +0.689 to +0.676, empirical +0.578 to +0.581) and the bucket AUCs move from 0.890 to 0.882 and from 0.833 to 0.824.

We knew that ICLR 2022--23 review data is likely inside every judge model's training data, and thought it possible that a judge model like the one employed by RAG-Novelty was matching the labels by recognizing the papers in question. To probe this, we ask \texttt{gpt-5-mini} and \texttt{gpt-4o-mini} to estimate the probability of acceptance from the verbatim title and abstract, from a deep paraphrase, and from the title alone, and we separately ask each model to name the venue on a 150-paper subset. While the models can still predict acceptance after paraphrase, the venue recall is near zero (Table~\ref{tab:memorization}).

We include all prompts below.

\begin{table}[t]
    \centering
    \begingroup
\small
\setlength{\tabcolsep}{3.5pt}
\begin{tabular}{lcccc}
\toprule
\rowcolor{gray!15}
 & \multicolumn{2}{c}{\textbf{Technical}} & \multicolumn{2}{c}{\textbf{Empirical}} \\
\rowcolor{gray!15}
\textbf{System} & $\boldsymbol{\rho}$ & \textbf{AUC} & $\boldsymbol{\rho}$ & \textbf{AUC} \\
\midrule
Yin & +0.06 & 0.52 & $-$0.03 & 0.46 \\
RND & $-$0.12 & 0.42 & $-$0.12 & 0.44 \\
SemNovel & +0.00 & 0.50 & $-$0.05 & 0.47 \\
FastTextLOF & $-$0.05 & 0.48 & $-$0.10 & 0.44 \\
NovaScore & +0.06 & 0.56 & +0.05 & 0.54 \\
AI-Sci Judge & +0.18$^{*}$ & 0.62 & +0.14 & 0.56 \\
ARC & $-$0.04 & 0.47 & $-$0.20$^{*}$ & 0.42 \\
\textbf{RAG-Novelty} & \textbf{+0.69$^{**}$} & \textbf{0.89} & \textbf{+0.58$^{**}$} & \textbf{0.83} \\
\bottomrule
\end{tabular}
\endgroup
    \caption{Spearman $\rho$ (with $p$) and high-vs-low ROC AUC of the eight core systems against ICLR 2022--23 reviewer novelty scores, per dimension ($n=200$ each). $^{*}p<.05$, $^{**}p<.001$.}
    \label{tab:iclr}
\end{table}

\begin{table}[H]
\centering
\begingroup
\small
\setlength{\tabcolsep}{4pt}
\begin{tabular}{lccccc}
\toprule
\rowcolor{gray!15}
\textbf{Model} & \textbf{verbatim} & \textbf{paraphr.} & \textbf{title} & \textbf{ICLR} & \textbf{+year} \\
\midrule
gpt-5-mini & 0.865 & 0.846 & 0.684 & 1.3\% & 0.7\% \\
gpt-4o-mini & 0.644 & 0.611 & 0.590 & 2.0\% & 0.7\% \\
\bottomrule
\end{tabular}
\endgroup
\caption{Memorization probe ($n=344$ ICLR 2022--23 papers): ROC AUC for predicting the true accept/reject decision from verbatim title+abstract, a deep paraphrase, and the title alone; and venue recall (does the model name ICLR / the correct year?) on a 150-paper subset.}
\label{tab:memorization}
\end{table}

\begin{myblock}{\textsc{Pool Keyword Prompt}}
    "You help simulate retrieval of prior related work for a research paper.
    Given the paper's title and abstract, propose short topical search queries
    that would surface prior art from the same research area on Semantic Scholar.
    Focus on the paper's problem space, methodology family, or key sub-field ---
    not terms so specific they only match the paper itself. Each query should be
    2--5 words."
\end{myblock}

\begin{myblock}{\textsc{Memorization Probe Prompts}}
    \textbf{Acceptance:} "The following paper was submitted to ICLR
    \{year\}. Based on everything you know, estimate the probability that it was
    ACCEPTED. \{content\} Respond with ONLY an integer 0-100 (your probability of
    acceptance in percent)."\\\\
    \textbf{Paraphrase (input form):} "Rewrite this abstract completely in your
    own words. Preserve every claim, method detail, and result, but change all
    sentence structures and wording so no phrase matches the original. Output
    only the rewritten abstract. \{abstract\}"\\\\
    \textbf{Venue recall:} "Below are the title and abstract of a research paper.
    If you recognize this specific paper, name the venue (conference/journal) and
    year where it was published or presented. If you do not recognize it, answer
    UNKNOWN. [...] Answer with just the venue name and year (e.g. 'NeurIPS 2021')
    or UNKNOWN."
\end{myblock}

\section{Taxonomy and System Selection}
\label{app:other_systems}
While there exist other novelty systems, we did not evaluate all of them. We explain the reasoning here. Note that although Si and OpenNovelty also output no scores, they do expose some judgment we can read out (\S\ref{sec:systems}); the first group below exposes nothing at all.

For the ``cite-only'' systems, we can roughly bucket them as (1) no separable novelty judgment to probe, such as AI-Researcher, Robin, and AI Scientist v2; (2) purely parametric judgments with no explicit pool we can manipulate, such as ResearchAgent, CycleReviewer, and SCI-Idea; and (3) scorers which use a very similar mechanism to a system already evaluated but are more expensive, such as Chain-of-Ideas (CoI) which is similar to the AI Scientist check and InnoEval which is similar to the OpenNovelty assessment. For DeepScientist, we choose not to evaluate this system because it assumes extensive iteration (thus the inclusion of its own findings in the pool) which does not match our model of novelty evaluation.

\begin{table*}[t]
\centering
\begingroup
\footnotesize
\setlength{\tabcolsep}{4pt}
\begin{tabular}{
  >{\raggedright\arraybackslash}p{2.6cm}
  >{\raggedright\arraybackslash}p{2.9cm}
  >{\raggedright\arraybackslash}p{3.1cm}
  >{\raggedright\arraybackslash}p{2.5cm}
  >{\raggedright\arraybackslash}p{3.3cm}}
\toprule
\rowcolor{gray!15}
\textbf{System} & \textbf{Output granularity} & \textbf{Pool observability} & \textbf{Comparison unit} & \textbf{Reference} \\
\midrule
\multicolumn{5}{l}{\emph{Evaluated --- full suite}} \\
Yin; RND; SemNovel & continuous & explicit corpus & document (embedding) & \citep{yin, rnd, semnovel} \\
FastTextLOF & continuous & explicit corpus & document (title embedding) & \citep{fasttext} \\
NovaScore & continuous & explicit corpus & claims & \citep{novascore} \\
RAG-Novelty & continuous & explicit corpus & document + retrieved context & \citep{lin2024evaluatingenhancinglargelanguage} \\
AI Scientist v1 check & binary ($\to$ logprob) & open retrieval ($\to$ sandboxed) & document & \citep{lu2024aiscientistfullyautomated} \\
ARC & binary & explicit corpus & document & \citep{liu2026autoresearchclawselfreinforcingautonomousresearch} \\
\addlinespace
\multicolumn{5}{l}{\emph{Evaluated --- scoped readouts (\S\ref{sec:systems})}} \\
Pairwise tournament & ranking & pool-blind (pairwise calls) & document pairs & \citep{si2024llmsgeneratenovelresearch} \\
OpenNovelty & free text + structured verdicts & open retrieval ($\to$ sandboxed) & claims $\times$ candidates & \citep{zhang2026opennoveltyllmpoweredagenticverifiable} \\
\addlinespace
\multicolumn{5}{l}{\emph{Surveyed (cite-only)}} \\
CoI novelty check & binary & open retrieval & idea & \citep{li2024chainideasrevolutionizingresearch} \\
InnoEval & ordinal & open retrieval & claims $\times$ evidence & \citep{qiao2026innoevalresearchideaevaluation} \\
ScholarEval & structured text & open retrieval & idea & \citep{moussa2026scholarevalresearchideaevaluation} \\
ResearchAgent reviewers & rubric scores & parametric & idea & \citep{baek2025researchagentiterativeresearchidea} \\
CycleReviewer & ordinal & parametric & document & \citep{weng2025cycleresearcherimprovingautomatedresearch} \\
DeepScientist & continuous & explicit corpus + own findings & idea & \citep{weng2025deepscientistadvancingfrontierpushingscientific} \\
SCI-IDEA & continuous & parametric & idea & \citep{keya2025sciideacontextawarescientificideation} \\
AI-Researcher & none (criterion in self-selection of own ideas) & --- & --- & \citep{tang2025airesearcherautonomousscientificinnovation} \\
Robin & none (fused into aggregate ranking) & --- & --- & \citep{ghareeb2025robinmultiagentautomatingscientific} \\
AI Scientist v2 & none (dissolved into ideation) & --- & --- & \citep{yamada2025aiscientistv2workshoplevelautomated} \\
\bottomrule
\end{tabular}
\endgroup

\caption{Taxonomy of surveyed systems along the three axes of \S\ref{sec:taxonomy}. Each row categorizes the novelty \emph{judgment} embedded in the system, not the surrounding pipeline. Free-text output is treated as having no granularity for the purposes of categorization. Arrows ($\to$) mark adaptations we made for evaluation (\S\ref{sec:systems}). The evaluated systems are chosen to span the cells as broadly as possible; the two systems whose outputs do not admit score orderings are evaluated through scoped adapted readouts (\S\ref{sec:systems}). \emph{Parametric} pool observability means the judgment draws only on the LLM's internal knowledge. \emph{None} means no separable novelty judgment exists. The classification is by design, not behavior; RAG-Novelty's score in practice ignores its retrieved context (\S\ref{sec:motivating}).}
\label{tab:taxonomy}
\end{table*}

\section{Probe Construction}
\label{app:probes}

\paragraph{Chance floors.} The binomial nulls are 50\%, 25\%, and 12.5\% for pairwise, two-step, and three-step orderings, respectively. These floors are conservative, as fully random scores satisfy the orderings less often.

\paragraph{Skip thresholds and time filters.} The V-far and V-near pools are time-filtered to papers published before the focal paper. We skip these probes when fewer than 100 papers remain after filtering. V-cited and V-primacy are similarly skipped for focal papers with fewer than 20 resolved references. However, we do not actively select for focal papers with many references, to avoid biasing our set. The T-accumulation windows are skipped when the base pool holds fewer than 400 papers ($W_i \geq 100$), and the T-subsumption slice is disjoint from the base pool by construction.

\paragraph{Task selection detail.} Task titles in PapersWithCode contain duplicates arising from minor naming variations, such as link prediction and link-prediction; we prompt an LLM to identify likely duplicates, then manually verify the proposed merges before consolidating. Among specific tasks we prefer, for instance, a task such as \textit{link prediction} over \textit{machine learning}. The three domains were chosen for the high availability of papers in the archive. Extending to non-AI domains is desirable but constrained by data availability; PapersWithCode is to our knowledge the only large tagged corpus of this kind, whereas the subject-area tagging in, for instance, ChemRxiv and EconPapers is too coarse.

\begin{table}[t]
\centering
\begingroup
\small
\setlength{\tabcolsep}{4pt}
\begin{tabular}{llr}
\toprule
\rowcolor{gray!15}
\textbf{Task} & \textbf{Distant task} & \textbf{Pool} \\
\midrule
\multicolumn{3}{l}{\emph{NLP}} \\
Code generation            & Drug discovery    & 1555 \\
Nat.\ lang.\ understanding & Optical flow est. & 1560 \\
Hallucination              & Optical flow est. & 1655 \\
\addlinespace
\multicolumn{3}{l}{\emph{CV}} \\
Scene understanding        & Drug discovery    & 1462 \\
Optical flow estimation    & Hallucination     & 1862 \\
Novel view synthesis       & EEG               & 1281 \\
3D object detection        & Drug discovery    & 1371 \\
\addlinespace
\multicolumn{3}{l}{\emph{Biomed}} \\
Drug discovery             & Optical flow est. & 1165 \\
Medical image analysis     & Code generation   & 1202 \\
EEG                        & Code generation   & 1432 \\
\bottomrule
\end{tabular}
\endgroup
\caption{Tasks by domain with each task's cross-domain distant task (V-far) and pool size.}
\label{tab:tasks}
\end{table}

\paragraph{Pool construction.} All probe pools are built by \texttt{axioms/construct\_pools.py}. For the redundancy axiom, R-exact adds a verbatim copy of the focal paper to the base pool, R-paraphrase adds the cached \texttt{gpt-5-mini} rephrase, and R-partial plants atomic claims extracted by \texttt{precompute\_claims.py} (\texttt{gpt-5-nano}, roughly seven claims per paper). Claims are extracted from the title and abstract, and claim $i$ is planted into the $i$-th most TF-IDF-similar pre-focal pool paper, appended as sentences to the end of the host paper's abstract. The main gradient uses the paraphrased claims, the R-verbatim contrast uses the original wording, and R-concentrated plants all claims into a single paper. Synthetic paper ids are focal-scoped so that manipulated texts never collide across focal papers, and NovaScore reads ACUs composed from the host paper's cached units plus the planted claims. For relevance, the far pool substitutes the cross-domain distant task of Table~\ref{tab:tasks} and the near pool the same-domain neighbor task, both time-filtered to pre-focal papers; the citation probes remove the focal paper's resolved references from the base pool or retain only them. For temporality, the base pool is sorted by date and cut into four equal-count windows, and the subsumption slice takes the oldest post-focal papers at the same count as one window, disjoint from the base pool by construction. Pass conditions are checked in \texttt{axioms/evaluate.py}.

The OpenNovelty attribution evaluation reuses these manipulations as planted pool papers, an exact duplicate with verbatim abstract and perturbed title so the plant is not aliased with the focal paper itself, the cached deep paraphrase, and the distributed claim plants, alongside unmanipulated control pools.

FastTextLOF reads only titles, so the claim-planting probes, which modify abstracts, are undefined for it and left blank in Table~\ref{tab:main}.

\paragraph{Prompts.} We give the rephrase prompt (R-paraphrase, \texttt{gpt-5-mini}) below with rephrase quality statistics across all ten tasks (Table~\ref{tab:rephrase-quality}), followed by the R-partial claim extraction and claim paraphrase prompts (both \texttt{gpt-5-nano}, one call each per focal paper).

\begin{myblock}{\textsc{Rephrasing Prompt}}
    "You are given a research paper excerpt. Your task is deep paraphrase — not surface
    substitution.\\\\
    Process:\\
    1. Internalize the full meaning: key claims, methods, results, and reasoning.\\
    2. Set aside the original text mentally.\\
    3. Rewrite from scratch in your own voice, as a knowledgeable author.\\\\
    Requirements:\\
    - Actively restructure sentences and paragraphs\\
    - Use entirely different vocabulary and syntax\\
    - No phrase over 4 words should match the original\\
    - Flip voice (active $<->$ passive) where natural\\
    - Technical proper nouns may remain unchanged\\\\
    Output only the rewritten text."
\end{myblock}

\begin{table}[H]
\centering
\begingroup
\small
\setlength{\tabcolsep}{3.5pt}
\begin{tabular}{lcccc}
\toprule
\rowcolor{gray!15}
\textbf{Task} & \textbf{R-1} & \textbf{R-2} & \textbf{R-L} & \textbf{Cos.} \\
\midrule
Code generation          & 0.569 & 0.227 & 0.436 & 0.895 \\
NL understanding         & 0.559 & 0.218 & 0.426 & 0.902 \\
Hallucination            & 0.563 & 0.218 & 0.426 & 0.900 \\
Scene understanding      & 0.577 & 0.236 & 0.446 & 0.913 \\
Optical flow est.        & 0.568 & 0.225 & 0.432 & 0.904 \\
Novel view synthesis     & 0.588 & 0.246 & 0.457 & 0.916 \\
3D object detection      & 0.575 & 0.233 & 0.441 & 0.910 \\
Drug discovery           & 0.566 & 0.221 & 0.436 & 0.914 \\
Medical image analysis   & 0.558 & 0.215 & 0.418 & 0.895 \\
EEG                      & 0.552 & 0.206 & 0.410 & 0.883 \\
\midrule
All ($n=1000$)           & 0.568 & 0.224 & 0.433 & 0.903 \\
\bottomrule
\end{tabular}
\endgroup

\caption{Statistics of generated (model: \texttt{gpt-5-mini}) title+abstract rephrases across all 10 tasks ($n=100$ per task): ROUGE-1/2/L (R-1/2/L) against the original and embedding cosine similarity (Cos.). An independent embedder (all-mpnet, outside the evaluated embedding family) gives mean cosine 0.93, so the semantic preservation is not an artifact of the evaluated embedders.}
\label{tab:rephrase-quality}
\end{table}

\begin{myblock}{\textsc{Claim Extraction Prompt}}
    "You are given the title and abstract of a research paper. Extract its
    atomic claims: the individual self-contained statements of contribution,
    method, finding, or result that the paper asserts.\\\\
    Requirements:\\
    - Target 4-10 claims (fewer only if the abstract truly contains fewer).\\
    - Each claim is ONE standalone declarative sentence, understandable in
    isolation (no dangling pronouns --- name the paper's method/setting).\\
    - Preserve technical proper nouns, dataset names, and numbers exactly.\\
    - Do not invent content absent from the text.\\\\
    Return ONLY a JSON object: \{"claims": ["claim 1", "claim 2", ...]\}"
\end{myblock}

\begin{myblock}{\textsc{Claim Paraphrase Prompt}}
    "You are given a JSON list of claims from a research paper. Your task is a
    deep paraphrase of EACH claim --- not surface substitution.\\\\
    For every claim: internalize its full meaning, set the original wording
    aside, and rewrite it from scratch with entirely different vocabulary,
    syntax, and sentence structure. No phrase over 4 words may match the
    original. Technical proper nouns, dataset names, and numbers may remain
    unchanged. The meaning must be fully preserved.\\\\
    Return ONLY a JSON object with the paraphrases in the SAME ORDER and SAME
    COUNT as the input list: \{"paraphrased": ["...", "..."]\}"
\end{myblock}

\section{In-the-Wild Pair Mining}
\label{app:inthewild}

We draw again upon the full review texts for 8{,}294 ICLR 2022--23 papers. This produces some 64{,}000 review sentences that mention a citation and which therefore become a candidate for our mining of $(A,Y)$ pairs where a reviewer says $A$ is not novel because of prior work $Y$. We use a simple regex prefilter to identify sentences which contain snippets like ``not novel,'' ``already been done,'' and ``very similar to,'' which results in 4{,}109 sentences. 

Next, we use a \texttt{gpt-5-nano} call and the prompt below to extract whether or not there is a novelty dispute, information about the prior work, and the type of dispute. We use the extracted information to attempt to find the arXiv id of the prior work $Y$. This produces 563 pairs, of which 353 involve a $Y$ which is not cited by $A$. Of these 563, we keep the pairs whose $Y$ is already present in our benchmark corpora, for a final 197 unique $(A,Y)$ pairs. Note that the matched pairs skew toward well-cited prior art, since both the corpora and reviewer mentions favor heavily cited papers. In theory, this should also be an easier form of the real-world problem setting.

A blind audit of 50 randomly sampled scored pairs, by an LLM given no project context, judged 90\% to be genuine novelty disputes and 94\% to link the correct prior work (86\% both).

We give the extraction prompt below.

\begin{myblock}{\textsc{Review-Snippet Extraction Prompt}}
    "You analyze snippets from ICLR peer reviews. Decide whether the reviewer
    is DISPUTING the novelty of the paper under review by pointing to specific
    prior work (i.e., claiming the paper's idea/method/result already exists,
    is too similar to, or is only an incremental change over cited prior work).
    Neutral citation mentions, requests to cite, or praise of novelty are NOT
    disputes. Return ONLY a JSON object with this exact schema:
    \begin{flushleft}\ttfamily\footnotesize
    \{"novelty\_dispute": true|false,\\
    \hspace*{1em}"confidence": 0.0-1.0,\\
    \hspace*{1em}"prior\_work": [\{"title": ..., "authors": ..., "year": ..., "arxiv\_id": ..., "how\_referenced": "verbatim phrase the reviewer used to refer to this work"\}],\\
    \hspace*{1em}"dispute\_type": "duplicate" | "substantial\_overlap" | "incremental" | "other"\}
    \end{flushleft}
    prior\_work must list only works the reviewer invokes AGAINST the paper's
    novelty (empty list if none, or if novelty\_dispute is false).
    dispute\_type: duplicate = reviewer says the same thing was already done;
    substantial\_overlap = core contribution heavily overlaps prior work;
    incremental = minor delta over prior work; other = anything else."
\end{myblock}

\section{Probe Premises}
\label{app:premises}

\paragraph{Axiom R.} We assume that the planted claims constitute coverage of $P$'s content. We extract these claims from $P$ itself and show that the paraphrased forms preserve the content in Appendix~\ref{app:probes}. We note that recombining known claims can result in novelty \citep{zhao2025novelty}; however, R-concentrated just requires recombination to outscore restatement, and R-partial only asks that a paper with covered claims score below the \textit{same paper} with none covered. The first inequality of R-concentrated reads redundancy at the document level. Because a strictly claim-level system (like NovaScore) may equate the concentrated and distributed pools, we read the contrast in behavior as a diagnostic of a system's comparison unit. Note also that R-concentrated differs from R-paraphrase. The paraphrase probe is a standalone restatement of $P$, while the concentrated plant buries $P$'s claims inside another paper's own content, a more realistic and therefore harder version of the same coverage.

Finally, one may ask whether claim-level probes apply to idea-stage pitches, which several evaluated pipelines score before any experiment exists. The extracted claims are largely contribution and method statements of the form ``we do $X$'' rather than empirical outcomes.

\paragraph{Axiom V (distance).} We assume that the distant-task pool is less relevant to $P$ than $P$'s own pool. To make this premise more reliable here, we measure and find that 0.06\% of focal citations resolve into distant pools, and only 0.7\% of focal papers are semantically closer to the distant pool than to their own task pool. In cases such as that of interdisciplinary transfer, this distance proxy is invalid because the ``distant'' field is in fact relevant. On the other hand, because we care about the aggregate performance of metrics, these cases, which are rare by definition, should not affect the overall evaluation of any given system. Moreover, a system that fails V-far by correctly detecting a shared mechanism across two fields would actually be performing a much harder kind of matching than the same-field paraphrase detection test, which indeed many of the tested systems fail.

\paragraph{Axiom V (citations).} We assume that cited papers are relevant prior work, a sort of semantic space understood by the authors. Although citation practices are often noisy \citep{tahamtan2018creativitysciencelinkcited}, our V-cited probe requires only the citations being on the whole very relevant.

\paragraph{Axiom T.} We assume that the windows differ only in temporal identity, which we enforce by using equal-count windows. We also assume for subsumption that later work will actually absorb $P$'s contribution, which we expect is true in aggregate but may not be true for a given paper. We measure this premise. The future slice spans a median of 232 days past the focal paper, and for 22\% of focal papers it contains at least one paper that cites the focal paper, a lower bound on absorption since parallel work may absorb ideas without citing. Restricting to these verified-absorption papers ($n{=}87$ per full-coverage system), pass rates against the newest window remain at or below chance for every system (Yin 0.48, RND 0.51, SemNovel 0.52, FastTextLOF 0.47, NovaScore 0.28, RAG-Novelty 0.24, ARC 0.02; the judge reaches 0.58 on $n{=}24$), matching the full-sample result.

\section{Scale and Attribution}
\label{app:attribution}

We do not exempt particular systems based on design and try to run every probe on every system where possible. While we consider robustness to the pool part of a novelty metric's performance, we run fixed-frame and size-matched runs in order to better attribute failures. SemNovel, for instance, was published with parameter $K{=}1000$ neighbors over 36M documents, which is clearly much larger compared to our pools and our K. FastTextLOF too assumes a larger pool, although of a lesser magnitude than that of SemNovel's. Therefore, we attempt to explain whether it is this difference in scale which causes poor performance or an issue with the method.

In order to do so, we build a fixed-frame control (\texttt{axioms/fixed\_frame\_control.py}). We fit a single t-SNE for SemNovel and a single fastText model for FastTextLOF on the union of the base, distant, near, and future pools plus the focal paper itself, once per focal paper. We hold a fixed $K=20$ and re-score each probe, changing only which papers count as candidates. We then compare the pass rates paired per focal paper against the per-pool refit (see main results). If a failure flips to a pass under the frozen frame, we consider this ``frame-driven,'' but if the metric fails in both settings we consider this method-driven.

Across tasks, SemNovel's far-distant pass rate rises from 0--40\% to 99--100\% under the frozen frame and the near rate from 0.5--50\% to 82--98\%, so those failures can be attributed to the frame. On the other hand, the subsumption probe flips from a relative success to relative failure, falling from 73--84\% to 15--37\%, which suggests the moving frame was helping there; we report both directions. In the wild, SemNovel's per-pool refit makes exact ties impossible and produces deltas of essentially zero mean against noise, 53\% of them wrong-way, which is why we exclude it from Table~\ref{tab:inthewild}.

FastTextLOF shows a similar pattern, although more mildly. Its far and near rates rise from 54--81\% to 84--99\% and from 50--71\% to 62--92\% under the frozen frame. We therefore conclude its relevance failures are also largely frame-driven. Its paraphrase detection barely moves (45--56\% to 50--68\%), however, and its temporal-window checks fail in both settings, so we conclude those weaknesses are method-driven, and its subsumption rate falls under the frozen frame just as SemNovel's does.
\section{Full Result Tables}
\label{app:tables}

\begin{table*}[t]
\centering
\small
\begingroup
\small
\setlength{\tabcolsep}{4pt}
\begin{tabular}{lrrrrrrr}
\toprule
\rowcolor{gray!15}
\textbf{Metric} & \textbf{base} & \textbf{25\%} & \textbf{50\%} & \textbf{75\%} & \textbf{100\%} & \textbf{verbatim} & \textbf{concentr.} \\
\midrule
Yin & 0.230 & 0.211 & 0.209 & 0.206 & 0.206 & 0.196 & 0.187 \\
RND & 70.185 & 69.121 & 68.561 & 67.814 & 67.364 & 66.099 & 68.823 \\
SemNovel & 204.187 & 205.300 & 206.376 & 205.130 & 206.724 & 211.409 & 206.078 \\
NovaScore & 0.642 & 0.617 & 0.610 & 0.606 & 0.593 & 0.576 & 0.594 \\
ARC & 0.992 & 0.992 & 0.992 & 0.992 & 0.992 & 0.965 & 0.963 \\
\bottomrule
\end{tabular}
\endgroup
\caption{R-partial coverage gradient: mean score in each metric's \emph{native} scale (only within-row comparisons are meaningful) as planted coverage rises base $\to$ 25/50/75/100\%, plus the verbatim and concentrated contrasts. Normalized view in Figure~\ref{fig:rp-gradient}.}
\label{tab:rp-gradient}
\end{table*}

\begin{table}[t]
\centering
\small
\begingroup
\small
\setlength{\tabcolsep}{4pt}
\begin{tabular}{lcc}
\toprule
\rowcolor{gray!15}
\textbf{Metric} & \textbf{R-partial gradient} & \textbf{Temporal windows} \\
\midrule
Yin & 10/10 & 10/10 \\
RND & 10/10 & 9/10 \\
SemNovel & 0/10 & 7/10 \\
FastTextLOF & — & 0/10 \\
NovaScore & 9/10 & 5/10 \\
AI-Sci Judge & 2/3 & 3/3 \\
RAG-Novelty & 1/10 & 9/10 \\
ARC & 0/10 & 0/10 \\
\bottomrule
\end{tabular}
\endgroup
\caption{Page trend test per metric: tasks significant at $\alpha=.05$ / tasks tested, for the R-partial gradient and the temporal windows.}
\label{tab:page}
\end{table}

\begin{table*}[t]
\centering
\small
\begingroup
\small
\setlength{\tabcolsep}{4pt}
\begin{tabular}{lrrrrrrr}
\toprule
\rowcolor{gray!15}
\textbf{Task} & \multicolumn{4}{c}{\textbf{mean focal rank}} & \multicolumn{3}{c}{\textbf{rank-drop rate}} \\
\rowcolor{gray!15}
 & base & +copy & +para. & +unrel. & copy & para. & unrel. \\
\midrule
Code generation & 11.94 & 12.99 & 13.46 & 13.61 & 0.71 & 0.68 & 0.73 \\
Optical flow & 12.21 & 13.04 & 13.73 & 13.24 & 0.64 & 0.69 & 0.66 \\
Drug discovery & 12.54 & 13.49 & 13.55 & 13.89 & 0.72 & 0.70 & 0.77 \\
\bottomrule
\end{tabular}
\endgroup
\caption{Si tournament control: mean focal rank and rank-drop rate under an inserted copy, paraphrase, or deterministic \emph{unrelated} pool paper.}
\label{tab:si-control}
\end{table*}

We summarize all significance statistics in Table~\ref{tab:per-task-sig}.

In Table~\ref{tab:main} we count ties as failures. However, since systems differ in output granularity, we additionally provide Table~\ref{tab:ties}, which decomposes every pairwise probe into pass, exact-tie, and wrong-way percentages, with two-sided sign tests on the non-tie movements.

Additionally, because focal papers within a task share overlapping pools, we provide Table~\ref{tab:per-task-sig}, which gives the per-task view robust to that dependence. LLM systems are scored with one cached call per instance, so near-chance rates carry sampling noise.

\begin{table*}[t]
\centering
\begingroup
\footnotesize
\setlength{\tabcolsep}{2.5pt}
\begin{tabular}{lcccccccc}
\toprule
\rowcolor{gray!15}
\textbf{Probe} & \textbf{Yin} & \textbf{RND} & \textbf{SemNovel} & \textbf{FastTextLOF} & \textbf{NovaScore} & \textbf{AI-Sci Judge} & \textbf{RAG-Novelty} & \textbf{ARC} \\
\midrule
R-exact & 100/0/0$^{*}$ & 100/0/0$^{*}$ & 43/0/57$^{\dagger}$ & 52/0/48 & 73/13/14$^{*}$ & 70/0/30$^{*}$ & 28/53/19$^{*}$ & 100/0/0$^{*}$ \\
R-paraphrase & 99/1/0$^{*}$ & 97/3/0$^{*}$ & 40/0/60$^{\dagger}$ & 50/0/50 & 49/27/24$^{*}$ & 73/0/27$^{*}$ & 26/54/19$^{*}$ & 60/40/0$^{*}$ \\
R-verbatim & 91/9/0$^{*}$ & 88/11/1$^{*}$ & 46/0/54$^{\dagger}$ & --- & 56/24/20$^{*}$ & 50/1/49 & 25/57/18$^{*}$ & 12/88/0$^{*}$ \\
R-partial (25\%) & 66/34/1$^{*}$ & 53/44/3$^{*}$ & 51/0/49 & --- & 43/32/25$^{*}$ & 50/0/49 & 24/57/19$^{*}$ & 2/98/1 \\
R-conc.\ $<$ base & 79/21/0$^{*}$ & 60/38/1$^{*}$ & 46/0/54$^{\dagger}$ & --- & 51/27/22$^{*}$ & 59/0/41$^{*}$ & 23/56/21 & 12/87/1$^{*}$ \\
R-conc.\ $<$ dist. & 69/8/23$^{*}$ & 7/32/60$^{\dagger}$ & 49/0/51 & --- & 35/30/35 & 54/0/46 & 18/58/24$^{\dagger}$ & 12/86/2$^{*}$ \\
V-far & 100/0/0$^{*}$ & 99/1/0$^{*}$ & 6/0/94$^{\dagger}$ & 65/0/35$^{*}$ & 45/36/19$^{*}$ & 90/0/10$^{*}$ & 29/55/16$^{*}$ & 2/98/0$^{*}$ \\
V-near & 98/1/2$^{*}$ & 93/2/5$^{*}$ & 2/0/98$^{\dagger}$ & 59/0/41$^{*}$ & 43/35/22$^{*}$ & 85/1/15$^{*}$ & 29/56/15$^{*}$ & 2/98/0$^{*}$ \\
V-graded & 91/0/9$^{*}$ & 63/31/6$^{*}$ & 50/0/50 & 56/0/44$^{*}$ & 32/46/21$^{*}$ & 60/1/39$^{*}$ & 19/60/21 & 0/100/0 \\
V-cited & 58/42/0$^{*}$ & 79/9/12$^{*}$ & 42/0/58$^{\dagger}$ & 45/0/55$^{\dagger}$ & 37/40/24$^{*}$ & 64/0/36$^{*}$ & 23/58/19 & 3/97/0$^{*}$ \\
V-primacy & 59/0/41$^{*}$ & 98/0/2$^{*}$ & 98/0/2$^{*}$ & 62/0/38$^{*}$ & 32/36/32 & 62/0/37$^{*}$ & 19/57/24 & 3/96/0$^{*}$ \\
T-acc.\ $W_1{>}W_2$ & 73/0/27$^{*}$ & 59/7/34$^{*}$ & 58/0/42$^{*}$ & 51/0/49 & 31/44/25$^{*}$ & 60/0/40$^{*}$ & 27/53/20$^{*}$ & 0/100/0 \\
T-acc.\ $W_2{>}W_3$ & 66/0/34$^{*}$ & 57/5/39$^{*}$ & 58/0/42$^{*}$ & 48/0/52 & 31/42/27 & 63/0/37$^{*}$ & 27/57/16$^{*}$ & 1/99/0$^{*}$ \\
T-acc.\ $W_3{>}W_4$ & 60/0/40$^{*}$ & 52/6/42$^{*}$ & 53/0/47$^{*}$ & 48/0/52 & 32/40/29 & 54/0/46 & 21/60/19 & 2/98/1 \\
T-sub.\ $<W_4$ & 42/0/58$^{\dagger}$ & 36/6/58$^{\dagger}$ & 55/0/45$^{*}$ & 50/0/50 & 31/38/31 & 40/0/60$^{\dagger}$ & 17/60/23$^{\dagger}$ & 2/97/1 \\
T-sub.\ $<$ base & 25/0/75$^{\dagger}$ & 57/4/38$^{*}$ & 100/0/0$^{*}$ & 54/0/46 & 29/33/38$^{\dagger}$ & 39/0/61$^{\dagger}$ & 20/57/23 & 2/96/2 \\
\bottomrule
\end{tabular}
\endgroup
\caption{Movement decomposition for every pairwise probe: percent pass\,/\,exact tie\,/\,wrong-way, pooled over tasks. $^{*}$non-tie movements favor the expected direction (two-sided sign test $p<.05$); $^{\dagger}$favor the wrong direction.}
\label{tab:ties}
\end{table*}

\begin{table}[t]
\centering
\small
\begingroup
\small
\setlength{\tabcolsep}{3.5pt}
\begin{tabular}{lrrrr}
\toprule
\rowcolor{gray!15}
\textbf{Metric} & \textbf{pass} & \textbf{tie} & \textbf{wrong} & \textbf{rate} \\
\midrule
Yin & 50 & 147 & 0 & 0.25 \\
RND & 67 & 126 & 4 & 0.34 \\
NovaScore$^{\ddagger}$ & 47 & 103 & 47 & 0.24 \\
\bottomrule
\end{tabular}
\endgroup
\caption{In-the-wild ICLR reviewer-named prior art: pass ($\Delta<0$), exact-tie, and wrong-way counts with pass rate, on the 197 unique (A, Y) pairs. $^{\ddagger}$The 47/47 split of non-ties (sign test $p \approx 1$) is likely noise.}
\label{tab:inthewild}
\end{table}

\begin{table*}[t]
\centering
\begingroup
\footnotesize
\setlength{\tabcolsep}{3.5pt}
\begin{tabular}{lcccccccc}
\toprule
\rowcolor{gray!15}
\textbf{Probe} & \textbf{Yin} & \textbf{RND} & \textbf{SemNovel} & \textbf{FastTextLOF} & \textbf{NovaScore} & \textbf{AI-Sci Judge} & \textbf{RAG-Novelty} & \textbf{ARC} \\
\midrule
\multicolumn{9}{l}{\emph{Axiom R --- Redundancy}} \\
R-exact & 10/10 & 10/10 & 0/10 & 2/10 & 10/10 & 3/3 & 0/10 & 10/10 \\
R-paraphrase & 10/10 & 10/10 & 0/10 & 0/10 & 1/10 & 3/3 & 0/10 & 8/10 \\
R-verbatim & 10/10 & 10/10 & 0/10 & — & 4/10 & 0/3 & 0/10 & 0/10 \\
R-partial (25\%) & 9/10 & 3/10 & 0/10 & — & 0/10 & 0/3 & 0/10 & 0/10 \\
R-partial gradient$^\dagger$ & 0/10 & 0/10 & 0/10 & — & 0/10 & 0/3 & 0/10 & 0/10 \\
R-concentrated $<$ base & 10/10 & 6/10 & 0/10 & — & 3/10 & 2/3 & 0/10 & 0/10 \\
R-concentrated $<$ distributed & 10/10 & 0/10 & 0/10 & — & 0/10 & 0/3 & 0/10 & 0/10 \\
\midrule
\multicolumn{9}{l}{\emph{Axiom V --- Relevance}} \\
V-far & 10/10 & 10/10 & 0/10 & 7/10 & 0/10 & 3/3 & 0/10 & 0/10 \\
V-near & 10/10 & 10/10 & 0/10 & 3/10 & 0/10 & 3/3 & 0/10 & 0/10 \\
V-graded (far $>$ near) & 10/10 & 5/10 & 5/10 & 3/10 & 0/10 & 2/3 & 0/10 & 0/10 \\
V-cited & 3/10 & 10/10 & 0/10 & 0/10 & 0/10 & 2/3 & 0/10 & 0/10 \\
V-primacy & 4/10 & 10/10 & 10/10 & 5/10 & 0/10 & 2/3 & 0/10 & 0/10 \\
\midrule
\multicolumn{9}{l}{\emph{Axiom T --- Temporality}} \\
T-acc.\ $W_1 > W_2$ & 10/10 & 5/10 & 5/10 & 1/10 & 0/10 & 2/3 & 0/10 & 0/10 \\
T-acc.\ $W_2 > W_3$ & 9/10 & 4/10 & 3/10 & 0/10 & 0/10 & 3/3 & 0/10 & 0/10 \\
T-acc.\ $W_3 > W_4$ & 8/10 & 3/10 & 3/10 & 0/10 & 0/10 & 1/3 & 0/10 & 0/10 \\
T-acc.\ monotone$^\dagger$ & 6/10 & 3/10 & 1/10 & 0/10 & 0/10 & 0/3 & 0/10 & 0/10 \\
T-sub.\ $< W_4$ & 0/10 & 0/10 & 1/10 & 0/10 & 0/10 & 0/3 & 0/10 & 0/10 \\
T-sub.\ $<$ base & 0/10 & 3/10 & 10/10 & 4/10 & 0/10 & 0/3 & 0/10 & 0/10 \\
\bottomrule
\end{tabular}
\endgroup
\caption{Per-task significance counts: for each probe and system, the number of tasks whose per-task pass rate individually exceeds chance (one-sided binomial, raw $p<.05$) over the number of tasks tested. Pooled Holm-corrected claims appear in Table~\ref{tab:main}; this view shows how uniformly each behavior holds across tasks. $^{\dagger}$chained checks (chance 0.125--0.25); $^{*}$3-task subset.}
\label{tab:per-task-sig}
\end{table*}

\section{Adapter Details}
\label{app:adapters}

Retrieval-based systems are sandboxed by redirecting their retrieval wholesale to the constructed pool, so the judge sees only pool papers; we do not compare against the deployed retrieval path.

Coarser output granularity shrinks the effect size of our statistical tests. A binary verdict, for instance, will only change if the movement is large enough to push across the threshold. A finer output granularity is therefore not only more interpretable but more useful for understanding the performance of the system, which motivates the log-probability readout below.

\paragraph{AI Scientist judge.} The prompts are taken verbatim from the released implementation and are not re-printed here. Our readout replaces the binary parse with the token log-probability $P(\text{novel})$, and retrieval is sandboxed to BM25 over the benchmark pool. We read the log-probability from an adjudication call we add after the loop. Across all 6{,}427 cached judge calls, thresholding this log-probability at 0.5 agrees with the native binary verdict 93.1\% of the time, indicating the relaxed readout is faithful to the judgment it relaxes.

\paragraph{NovaScore.} The ACU-extraction and NLI prompts are reproduced verbatim from the NovaScore paper's Appendix A.1 and are not re-printed here. Retrieval is SBERT with FAISS over pool ACUs.

\paragraph{ARC.} This check is purely lexical and there is no prompt. The gate scores $1 - \max\, (0.7 \cdot \text{keyword Jaccard} + 0.3 \cdot \text{title string similarity})$ over the five most similar pool papers, ignoring papers below similarity 0.25. We pass the focal title as its topic string, the title and abstract as its hypothesis text, and our pool as its seen-papers list, with web search disabled.

\paragraph{RAG-Novelty.} The prompt below is the single-paper form of the original released prompt, with the pairwise framing and its Step 3 removed and the year parameterized. Retrieval uses \texttt{text-embedding-3-small} with $k{=}10$. Also, rerunning RAG-Novelty under \texttt{gpt-4o-mini} on code generation reproduces the same tie-dominated profile (pass rates 0.10--0.21, ties 51--72\%), so its insensitivity is \textit{architectural} and should not be considered a property of the model.

\paragraph{Si tournament.} The prompt below is verbatim from the released tournament code; we follow the paper in that each tournament has 20 entrants (for us, the focal paper plus its 19 nearest pool papers by embedding) ranked over 5 Swiss-style rounds by \texttt{gpt-5-mini}, with entrants re-paired each round by accumulated score. The verdict we use is the focal paper's fractional rank under an inserted copy, paraphrase, or unrelated paper.

\paragraph{OpenNovelty.} OpenNovelty outputs free text with structured verdicts, so to adapt it to our evaluation, we parse the output for the planted paper ID, in its related-work classification or its cited refutation evidence. We also read the fraction of contributions the structured verdict marked as refuted.

\begin{myblock}{\textsc{RAG-Novelty Prompt (adapted, single-paper form)}}
    "You are an advanced language model tasked with determining the novelty of
    research papers in \{year\}. Your goal is to evaluate the novelty of a
    research paper based on its title and abstract.\\\\
    Step 1: Independent Evaluation\\
    Analyze the research paper's title and abstract independently.
    Similar abstracts have been retrieved from a vector database based on the
    provided abstract.
    Contextual Date Analysis: The average published date of the retrieved
    documents is provided. Use this average date as additional context for your
    evaluation.
    Consider that papers with an average date that is later or more recent in
    time are generally more novel.
    Consider the following aspects for the paper:
    Novelty of Methodology: Are the methods used new and innovative?
    Surprisingness of Findings: Are the findings unexpected or counterintuitive?
    Impact on Existing Knowledge: How does the research challenge or expand
    current scientific understanding?
    Potential for Future Research: Does the paper open up new directions for
    research?
    Relevance to \{year\} Scientific Understanding: How well does the paper align
    with or push the boundaries of current trends?\\\\
    Step 2: Quantitative Assessment\\
    Assign a score from 1-10 to the research paper for its novelty, with 10 being
    the most novel. This score should be based on the content of the title and
    abstract, as well as the contextual information from the average published
    date.
    Provide a brief justification for the score, using specific quotes and
    context.
    Return your final answer as a JSON object on the last line:
    \{"score": $<$1-10$>$\}"\\\\
    \textbf{Per-paper input:} "Paper Average Cosine Similarity (higher is
    better): \{avg\_sim\} | Paper Average Contextual Date (more recent is
    better): \{avg\_date\} | Paper Title: \{title\} | Paper Abstract: \{abstract\}"
\end{myblock}

\begin{myblock}{\textsc{Si Pairwise Tournament Prompt (verbatim)}}
    "You are a reviewer specialized in Natural Language Processing and Large
    Language Models. You are given two project summaries. One of them is
    accepted by a top AI conference (like ICLR or ACL) and the other one is
    rejected. Your task is to identify the one that has been accepted.\\
    The two project proposals are:\\\\
    paper 1:\\
    \{idea\_1\}\\\\
    paper 2:\\
    \{idea\_2\}\\\\
    Now decide which one is the accepted idea. Directly return a number 1 or 2
    and nothing else."
\end{myblock}

\section{Compute and Configuration}
\label{app:ledger}

\paragraph{Release and licenses.} We release only arXiv IDs, derived files, and scripts that fetch paper texts from the public arXiv and Semantic Scholar APIs, so no actual paper text is redistributed.

We get task tags from the CC BY-SA PapersWithCode archive. Third-party system implementations are not redistributed; our release pins and fetches their original repositories, with our adapters and any small patches applied on top. Our adapters do embed the systems' released prompts verbatim, and the tournament adapter is a port of the released ranking code; both are reproduced under the sources' open licenses (MIT, Apache-2.0) with attribution.

\paragraph{Compute disclosure.} All experiments, embedding computation (BGE-M3, SBERT, fastText), t-SNE, and analysis were run locally with no GPU cluster usage. API usage (OpenAI; \texttt{gpt-5-nano}, \texttt{gpt-5-mini}, \texttt{gpt-4o-mini}, and \texttt{text-embedding-3-small}) across all runs reported in this paper totaled just under \$400.

\paragraph{Per-system configuration.} For Yin, RND, and SemNovel, we embed titles and abstracts with BGE-M3; additionally for SemNovel we project with openTSNE and score mean K-NN distance in the 2D projection with $K$ set to $\max(10, 2\%)$ of the pool. For FastTextLOF, we train from scratch a fastText skipgram per pool on titles and score a local outlier factor. For NovaScore we use \texttt{gpt-5-nano} at minimal reasoning effort with SBERT retrieval over pool ACUs. For the AI Scientist judge we use \texttt{gpt-4o-mini} in logprob mode, while RAG-Novelty, the Si tournament, and OpenNovelty use \texttt{gpt-5-mini} (which does not have logprob mode). Sampling settings are API defaults throughout, as \texttt{gpt-5}-class models reject non-default values.

We evaluate the AI Scientist judge on a three-task subset with one task per domain (code generation, optical flow estimation, drug discovery). We do not evaluate it on all ten tasks because its search loop makes the API cost prohibitive. Similarly, we scope OpenNovelty's attribution evaluation to 50 focal papers on each of two tasks (drug discovery, code generation). This is because its agentic pipeline takes $\sim$0.5M tokens per record making it the most expensive per paper of any system we run.
\end{document}